%% file: main.tex
\documentclass[a4paper, 11pt]{article}
\pdfoutput=1  
\usepackage{etoolbox,verbatim}
\usepackage{latexml} % provides \iflatexml; false for pdfLaTeX

\newtoggle{arxiv}
\toggletrue{arxiv} 

\newtoggle{customthms}
\toggletrue{customthms} 

\newtoggle{colt}
\togglefalse{colt} 

\input{preamble/_preamble_includes}
\iflatexml % HTML: render wrapped figures as ordinary figures
\makeatletter
\expandafter\def\csname begin{wrapfigure}\endcsname{\@ifnextchar[{\fada@wrapopt}{\fada@wrap}}
\def\fada@wrapopt[#1]{\fada@wrap}
\def\fada@wrap#1#2{\begin{figure}[h]}
\expandafter\def\csname end{wrapfigure}\endcsname{\end{figure}}
\makeatother
\fi


\title{\textbf{FADA}: \textbf{F}ew-Shot Domain \textbf{A}daptation via \textbf{D}ynamics \textbf{A}lignment for Humanoid Control}

\iflatexml
\author{Angchen Xie\textsuperscript{*} \and Nikhil Sobanbabu\textsuperscript{*} \and Ishayu Shikhare \and Alan Wang \and Max Simchowitz \and Guanya Shi}
\makeatletter
\ifdefined\lx@@@contact
  \@add@frontmatter{ltx:creator}[role=author]{\lx@@@contact{institution}{Carnegie Mellon University \quad \textsuperscript{*}Equal contribution}}
\else
  \date{Carnegie Mellon University \quad \textsuperscript{*}Equal contribution}
\fi
\makeatother
\else
\author{
Angchen Xie\textsuperscript{*}\quad
Nikhil Sobanbabu\textsuperscript{*}\quad
Ishayu Shikhare\quad
Alan Wang
\\
Max Simchowitz\quad
Guanya Shi
\\
\vspace{0em}
\rule{.72\textwidth}{.7pt}
\\
\footnotesize
Carnegie Mellon University \quad \textsuperscript{*}Equal contribution
}
\fi
 
\begin{document}

\iflatexml
  \maketitle
  \input{sections/intro_fig}
\else
\begin{tcolorbox}[
    colback=steelgray, colframe=irongray,
    boxrule=0pt,
    arc=2mm
  ]
  \maketitle
  \tcbline
  \input{sections/intro_fig}
\end{tcolorbox}
\fi

\vspace{-1.0em}
\begin{abstract}\input{sections/abstract}
\end{abstract}

\input{sections/intro}

\input{sections/related}
\input{sections/framework_fig}
\input{sections/problem_setup}
\input{sections/method}

\input{sections/experiments}

\input{sections/conclusion_and_limitations}

\input{sections/acknowledgements}

\newpage
\newpage
\bibliographystyle{plainnat}
\bibliography{refs}
\newpage 
\appendix

\input{sections/appendix}

\end{document}

%% file: preamble/_preamble_includes.tex
\input{preamble/color_defs}

\input{preamble/header}
\input{preamble/theorem_styles}

\input{preamble/characters}
\input{preamble/general_macros}

\input{preamble/specific_macros}

\input{preamble/algnames}
\iftoggle{arxiv}
{
\iflatexml\else\input{preamble/title_format}\fi
}
{}

%% file: preamble/color_defs.tex
\usepackage[table]{xcolor}
\usepackage{tcolorbox}
\tcbuselibrary{skins,breakable}

\tcbset{
  aibox/.style={
      width=\linewidth,
      top=6pt,
      bottom=0pt,
      left=1.5pt,
      right=1.5pt,
      colback=blue!60!gray!5,
      colframe=black,
      colbacktitle=black,
      enhanced,
      center,
      attach boxed title to top left={yshift=-0.1in,xshift=0.15in},
      boxed title style={boxrule=0pt,colframe=white,},
    }
}
\newtcolorbox{AIbox}[2][]{aibox,title=#2,#1}

\definecolor{primalcolor}{HTML}{960000}
\definecolor{contrarycolor}{HTML}{009696}
\definecolor{primaldarkcolor}{HTML}{660000}
\definecolor{contrarydarkcolor}{HTML}{006666}
\definecolor{irongray}{HTML}{6D6E71}
\definecolor{steelgray}{HTML}{F1F4F7}
\definecolor{darkcontrarycolor}{HTML}{004C4C}
\definecolor{lightblue}{HTML}{2970CC}
\definecolor{lightpurple}{HTML}{673147}
\definecolor{ForestGreen}{HTML}{FF5733}
\definecolor{myred}{HTML}{AA4A44}
\definecolor{hyppurple}{HTML}{800080}

\newcommand{\linkcolor}{contrarydarkcolor}
\newcommand{\urlcolor}{contrarydarkcolor}
\newcommand{\citecolor}{contrarydarkcolor}

\newcommand{\thmcolordark}{primaldarkcolor}

\definecolor{verylightgray}{RGB}{240, 240, 240}
\definecolor{srcband}{RGB}{214, 233, 255}
\definecolor{tgtband}{RGB}{201, 201, 247}
\definecolor{realband}{RGB}{246, 208, 208}
\definecolor{hwband}{RGB}{220, 242, 220}
\definecolor{methodred}{HTML}{A60000}
\newcommand{\goodnumber}[1]{{\color{methodred}\textbf{#1}}}

%% file: preamble/header.tex
\usepackage[T1]{fontenc}
\usepackage{courier}
\usepackage{capt-of}
\usepackage[font=footnotesize,labelfont=bf]{caption}
\usepackage{wrapfig}

\iftoggle{arxiv}{\usepackage{natbib}}{}
\usepackage{scalefnt}
\usepackage{amssymb,upgreek,bm}
\usepackage{mathtools}
\usepackage[english]{babel}  
\usepackage{multirow,tcolorbox,tikz,tikz-cd,turnstile,xspace,xparse}
\usetikzlibrary{arrows.meta,positioning,fit,calc}
\usepackage{relsize,breakcites,appendix}
\usepackage{longtable,tablefootnote,booktabs,float,makecell}
\IfFileExists{algorithm.sty}{
    \usepackage{algorithm}
}{}
\usepackage[normalem]{ulem}
\usepackage{tabu}
\usepackage[shortlabels]{enumitem} 
\usepackage{footnote}
\usepackage{boxedminipage}
\usepackage{multirow,nicefrac}
\usepackage{verbatim}

\makeatletter
\@ifundefined{nolinenumbers}{\newenvironment{nolinenumbers}{}{}}{}
\makeatother

\makeatletter
\@ifpackageloaded{hyperref}{
  \hypersetup{colorlinks=true,linkcolor=\linkcolor,urlcolor=\urlcolor,citecolor=\citecolor,breaklinks}
}{
  \usepackage[colorlinks=true,linkcolor=\linkcolor,urlcolor=\urlcolor,citecolor=\citecolor,breaklinks]{hyperref}
}
\makeatother
\usepackage{prettyref}

\usepackage[capitalise,nameinlink]{cleveref}
\crefname{appendix}{Appendix}{Appendices}
\Crefname{appendix}{Appendix}{Appendices}

\iftoggle{colt}{
    \DeclareRobustCommand{\qed}{
        \usepackage{thmtools}
          \ifmmode \mathqed
          \else
            \leavevmode\unskip\penalty9999 \hbox{}\nobreak\hfill
            \quad\hbox{\qedsymbol}%
          \fi
    }
}
{
    \usepackage{amsthm}
     
}
\usepackage{thm-restate}

\iftoggle{arxiv}
{
\usepackage{mathrsfs}

    \usepackage[
        a4paper,
        top=1in,
        bottom=1in,
        left=1in,
        right=1in]{geometry}

    \iflatexml\else % LaTeXML (arXiv HTML) does not support these font packages
    \IfFileExists{mathdesign.sty}{
        \usepackage[bitstream-charter]{mathdesign}
        
    }{}
    \IfFileExists{PTSans.sty}{
        \usepackage[scaled=0.92]{PTSans}
    }{}
    \fi

    \usepackage{subfig}

    \usepackage{fullpage}
    \IfFileExists{algpseudocode.sty}{
        \usepackage[noend]{algpseudocode}
    }{}
    \IfFileExists{algorithm.sty}{
        \usepackage{algorithm}
    }{}
   
}
{
    \usepackage{mathrsfs}
}

\DeclareMathAlphabet{\mathbfsf}{\encodingdefault}{\sfdefault}{bx}{n}

\numberwithin{equation}{section}

\iftoggle{arxiv}
{
    }
{
    
}
\iftoggle{arxiv}
{
    
}
{
    
}

%% file: preamble/theorem_styles.tex
\usepackage{thmtools}

\Crefname{equation}{Eq.}{Eqs.}
\Crefname{assumption}{Assumption}{Assumptions}
\Crefname{condition}{Condition}{Conditions}
\Crefname{claim}{Claim}{Claims}
\Crefname{property}{Property}{Properties}
\Crefname{construction}{Construction}{Constructions}

\declaretheoremstyle[
    headformat=\normalfont\textcolor{\thmcolordark}{\bfseries\NAME\,\NUMBER}\NOTE,%
    notefont={\normalfont\textcolor{\thmcolordark}{\bfseries}}, 
    notebraces={}{},
    bodyfont=\normalfont\itshape,
    spaceabove = 6pt,
    spacebelow = 6pt,
    ]{coloredthmversion}

\declaretheoremstyle[
    headformat=\normalfont\textcolor{\thmcolordark}{\bfseries\NAME\,\NUMBER}\NOTE,%
    bodyfont=\normalfont\itshape,
    spaceabove = 6pt,
    spacebelow = 6pt,
    ]{coloredthm}

\declaretheoremstyle[
    headformat=\normalfont\textcolor{\thmcolordark}{\bfseries\NAME\,\NUMBER}\NOTE,%
    bodyfont=\normalfont,
    spaceabove = 6pt,
    spacebelow = 6pt,
    ]{coloreddef}

\iftoggle{customthms}
{
    \theoremstyle{coloredthmversion}
}
{}

\iftoggle{customthms}{
  \theoremstyle{coloredthm}
}
{
  \theoremstyle{plain}
}
\newtheorem{theorem}{Theorem}
\newtheorem{lemma}{Lemma}[section]
\newtheorem{corollary}{Corollary}[section]
\newtheorem{proposition}[lemma]{Proposition}

\newtheorem*{thminformal*}{Informal Theorem}

\iftoggle{customthms}{
    \theoremstyle{coloreddef}
}
{
  \theoremstyle{definition}
}
\newtheorem{definition}{Definition}[section]

\newtheorem{assumption}{Assumption}[section]
\newtheorem{condition}{Condition}[section]

\makeatletter
\newcommand{\neutralize}[1]{\expandafter\let\csname c@#1\endcsname\count@}
\makeatother

\iftoggle{customthms}{
    \newtheoremstyle{named}{}{}{\itshape}{}{\bfseries}{}{.5em}{\Cref{#3} {\normalfont (informal)} }{}
    \theoremstyle{named}
    
    \theoremstyle{plain}
}
{}

\newtheorem*{theorem*}{Theorem}
\newtheorem*{lemma*}{Lemma}
\newtheorem*{corollary*}{Corollary}
\newtheorem*{proposition*}{Proposition}
\newtheorem*{claim*}{Claim}
\newtheorem*{fact*}{Fact}
\newtheorem*{observation*}{Observation}
\newtheorem*{definition*}{Definition}
\newtheorem*{remark*}{Remark}
\newtheorem*{example*}{Example}

%% file: preamble/characters.tex
\def\ddefloop#1{\ifx\ddefloop#1\else\ddef{#1}\expandafter\ddefloop\fi}
\def\ddef#1{\expandafter\def\csname bb#1\endcsname{\ensuremath{\mathbb{#1}}}}
\ddefloop ABCDEFGHIJKLMNOPQRSTUVWXYZ\ddefloop

\def\ddefloop#1{\ifx\ddefloop#1\else\ddef{#1}\expandafter\ddefloop\fi}
\def\ddef#1{\expandafter\def\csname frak#1\endcsname{\ensuremath{\mathfrak{#1}}}}
\ddefloop ABCDEFGHIJKLMNOPQRSTUVWXYZ\ddefloop

\def\ddefloop#1{\ifx\ddefloop#1\else\ddef{#1}\expandafter\ddefloop\fi}
\def\ddef#1{\expandafter\def\csname fr#1\endcsname{\ensuremath{\mathfrak{#1}}}}
\ddefloop ABCDEFGHIJKLMNOPQRSTUVWXYZ\ddefloop

\def\ddefloop#1{\ifx\ddefloop#1\else\ddef{#1}\expandafter\ddefloop\fi}
\def\ddef#1{\expandafter\def\csname eul#1\endcsname{\ensuremath{\EuScript{#1}}}}
\ddefloop ABCDEFGHIJKLMNOPQRSTUVWXYZ\ddefloop

\def\ddefloop#1{\ifx\ddefloop#1\else\ddef{#1}\expandafter\ddefloop\fi}
\def\ddef#1{\expandafter\def\csname scr#1\endcsname{\ensuremath{\mathscr{#1}}}}
\ddefloop ABCDEFGHIJKLMNOPQRSTUVWXYZ\ddefloop

\def\ddefloop#1{\ifx\ddefloop#1\else\ddef{#1}\expandafter\ddefloop\fi}
\def\ddef#1{\expandafter\def\csname b#1\endcsname{\ensuremath{\mathbf{#1}}}}
\ddefloop ABCDEFGHIJKLMNOPQRSTUVWXYZ\ddefloop

\def\ddefloop#1{\ifx\ddefloop#1\else\ddef{#1}\expandafter\ddefloop\fi}
\def\ddef#1{\expandafter\def\csname bhat#1\endcsname{\ensuremath{\hat{\mathbf{#1}}}}}
\ddefloop ABCDEFGHIJKLMNOPQRSTUVWXYZ\ddefloop

\def\ddefloop#1{\ifx\ddefloop#1\else\ddef{#1}\expandafter\ddefloop\fi}
\def\ddef#1{\expandafter\def\csname btil#1\endcsname{\ensuremath{\tilde{\mathbf{#1}}}}}
\ddefloop ABCDEFGHIJKLMNOPQRSTUVWXYZ\ddefloop

\def\ddefloop#1{\ifx\ddefloop#1\else\ddef{#1}\expandafter\ddefloop\fi}
\def\ddef#1{\expandafter\def\csname bst#1\endcsname{\ensuremath{\mathbf{#1}^\star}}}
\ddefloop ABCDEFGHIJKLMNOPQRSTUVWXYZ\ddefloop

\def\ddefloop#1{\ifx\ddefloop#1\else\ddef{#1}\expandafter\ddefloop\fi}
\def\ddef#1{\expandafter\def\csname bst#1\endcsname{\ensuremath{\mathbf{#1}^\star}}}
\ddefloop abcdeghijklmnopqrstuvwxyz\ddefloop

\def\ddefloop#1{\ifx\ddefloop#1\else\ddef{#1}\expandafter\ddefloop\fi}
\def\ddef#1{\expandafter\def\csname bhat#1\endcsname{\ensuremath{\hat{\mathbf{#1}}}}}
\ddefloop abcdefghijklmnopqrstuvwxyz\ddefloop

\def\ddefloop#1{\ifx\ddefloop#1\else\ddef{#1}\expandafter\ddefloop\fi}
\def\ddef#1{\expandafter\def\csname b#1\endcsname{\ensuremath{\mathbf{#1}}}}
\ddefloop abcdeghijklnopqrstuvwxyz\ddefloop

\def\ddefloop#1{\ifx\ddefloop#1\else\ddef{#1}\expandafter\ddefloop\fi}
\def\ddef#1{\expandafter\def\csname barb#1\endcsname{\ensuremath{\bar{\mathbf{#1}}}}}
\ddefloop abcdefghijklmnopqrstuvwxyz\ddefloop

\def\ddef#1{\expandafter\def\csname c#1\endcsname{\ensuremath{\mathcal{#1}}}}
\ddefloop ABCDEFGHIJKLMNOPQRSTUVWXYZ\ddefloop
\def\ddef#1{\expandafter\def\csname h#1\endcsname{\ensuremath{\widehat{#1}}}}
\ddefloop ABCDEFGHIJKLMNOPQRSTUVWXYZ\ddefloop
\def\ddef#1{\expandafter\def\csname hc#1\endcsname{\ensuremath{\widehat{\mathcal{#1}}}}}
\ddefloop ABCDEFGHIJKLMNOPQRSTUVWXYZ\ddefloop
\def\ddef#1{\expandafter\def\csname t#1\endcsname{\ensuremath{\widetilde{#1}}}}
\ddefloop ABCDEFGHIJKLMNOPQRSTUVWXYZ\ddefloop
\def\ddef#1{\expandafter\def\csname tc#1\endcsname{\ensuremath{\widetilde{\mathcal{#1}}}}}
\ddefloop ABCDEFGHIJKLMNOPQRSTUVWXYZ\ddefloop

%% file: preamble/general_macros.tex
\usepackage{tcolorbox}
\tcbuselibrary{skins,breakable}

\tcbset{
  aibox/.style={
      width=\linewidth,
      top=6pt,
      bottom=0pt,
      left=1.5pt,
      right=1.5pt,
colback=blue!60!gray!5,
colframe=black,
      colbacktitle=black,
      enhanced,
      center,
      attach boxed title to top left={yshift=-0.1in,xshift=0.15in},
      boxed title style={boxrule=0pt,colframe=white,},
    }
}

\iftoggle{arxiv}
{

}
{

}

\iftoggle{arxiv}
{

}
{

}

\newcommand{\ballkr}[1][r]{\cB_{k}(r)}

\DeclareMathSymbol{\shortminus}{\mathbin}{AMSa}{"39}

%% file: preamble/specific_macros.tex
\newcommand{\idea}{FADA\xspace}

\newcommand{\copred}{TF--CoPred\xspace}
\newcommand{\tfdagger}{TF--DAgger\xspace}

\providecommand{\goodnumber}[1]{\textbf{#1}}

\newcommand{\srcdom}{\Omega_{\mathrm{src}}}

%% file: sections/intro_fig.tex
\begin{nolinenumbers}
\begin{figure}[H]
    \vspace{-1.0em}
    \centering
    \includegraphics[width=\linewidth]{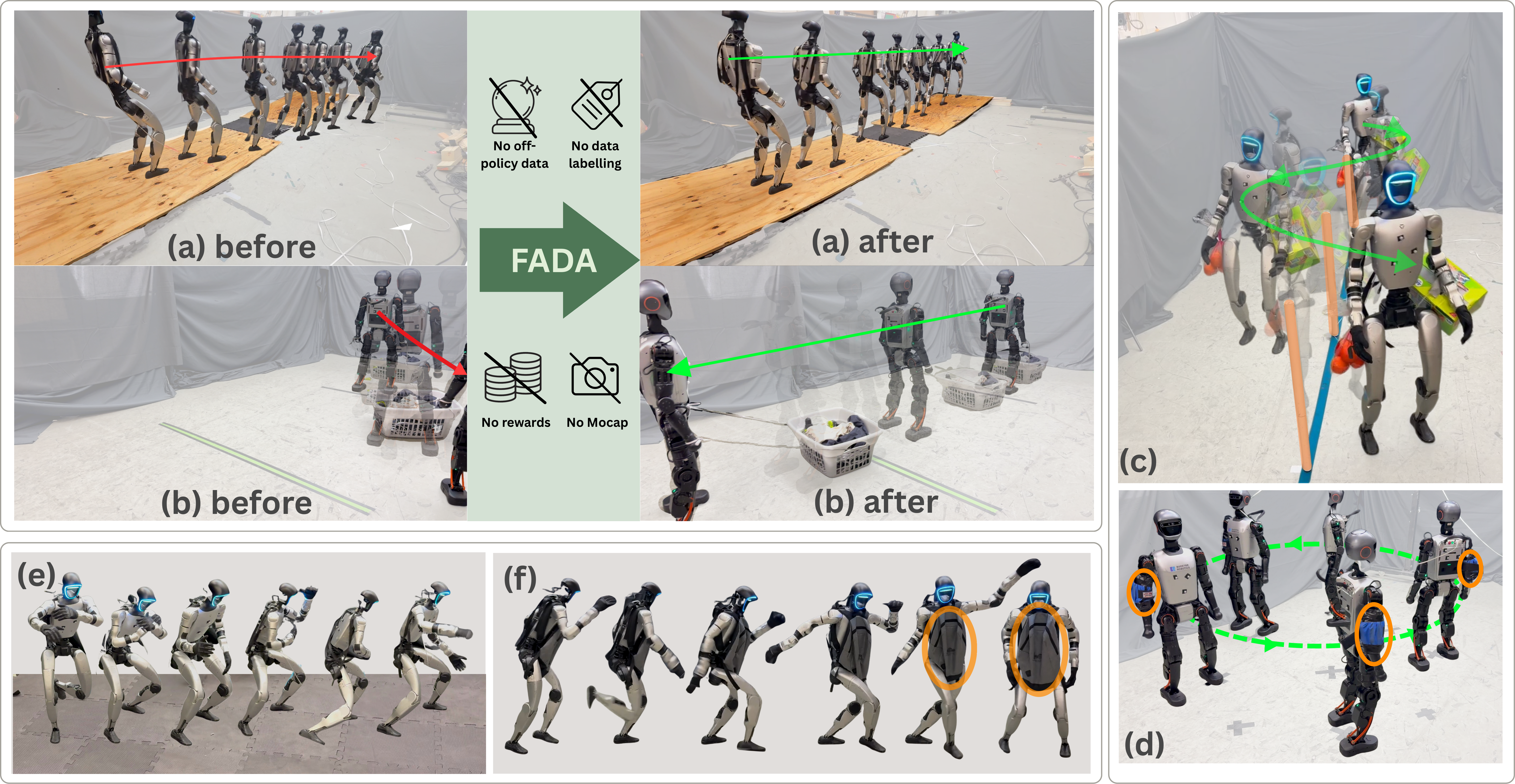}
    \caption{\scriptsize\textbf{\idea enables high-precision whole-body skills through dynamics alignment.}
    Through few-shot adaptation, humanoid robots can stably execute diverse real-world tasks that fail under zero-shot transfer.
    (a) and (b) illustrate the adaptation effect: Only after adaptation is Unitree G1 able to precisely track a line on a slope, and Booster T1 able to pull a 6~kg laundry basket across the finish line.
    (c)--(f) show additional adapted deployments: (c) G1 weaving through poles with asymmetric 3~kg groceries, (d) T1 tracking a circular path with an asymmetric 1~kg arm payload, (e) G1 executing Kung Fu motions on soft mats, and (f) G1 dancing with a 3.2~kg front payload.}
    \label{fig:teaser}
\end{figure}
\end{nolinenumbers}

%% file: sections/abstract.tex
High-precision humanoid control is limited by target-domain dynamics mismatch, where the same control objective can induce different realized motions under changes in terrain, payload, or actuator response. Existing methods either pursue zero-shot transfer through domain randomization or in-context adaptation without target-domain specialization, or require heavy adaptation pipelines that leverage target-domain data, such as model calibration, residual learning, or policy retraining. In this paper, we present \textbf{FADA} (\textbf{F}ew-Shot Domain \textbf{A}daptation via \textbf{D}ynamics \textbf{A}lignment), a three-stage Planner--Inverse Dynamics Model (Planner--IDM) framework for few-shot adaptation in humanoid control. \textbf{FADA} first trains an oracle policy with privileged information and then distills the oracle behavior into a deployable Planner--IDM student through DAgger. At deployment, \textbf{FADA} freezes the planner and finetunes only the IDM using approximately 2 minutes of target-domain rollouts with standard supervised learning. Rather than requiring optimal demonstrations or rewards, \textbf{FADA} uses the paired actions and observations that are observed during these rollouts as supervision, aligning the IDM's action generation with target-domain dynamics. Experiments show that \textbf{FADA} outperforms both in-context and end-to-end adaptation baselines, improving task performance under dynamics shifts and enabling real humanoid robots to execute diverse high-precision whole-body tasks.
Implementation details and qualitative hardware rollout videos are available at \url{https://lecar-lab.github.io/FADA-humanoid/}.

%% file: sections/intro.tex
\section{Introduction}

Humanoid robots must perform precise simultaneous locomotion and manipulation in complex environments.
Recent reinforcement learning methods have demonstrated agile and robust locomotion~\citep{gu2024humanoidgym}, whole-body tracking~\citep{he_asap_2025,he2024hover}, teleoperation~\citep{he2024omnih2o,fu2024humanplus}, and loco-manipulation~\citep{zhao_resmimic_2025,zhang2025falcon}.
These advances rely heavily on simulation, where policies can safely experience falls, impacts, contacts, and task perturbations at scale.
However, despite success in simulation,  these methods still struggle with \emph{precise} control on real hardware. When deployment conditions differ from training, the same control intent may no longer produce the same body motion.
This gap arises from a dynamics mismatch between simulation and the target domain, including differences in terrain, payload, and actuator response.
For humanoids, these discrepancies are amplified by tightly coupled whole-body dynamics: small errors in foot placement, contact timing, or force delivery can lead to posture drift, degraded tracking, or instability.
Such failures often do not indicate an unreasonable high-level command; rather, they arise when the motion or control output no longer induces the intended physical response in the target domain.

\begin{wrapfigure}{r}{0.43\linewidth}
    \vspace{-1.0em}
    \centering
    \iflatexml\includegraphics[width=0.6\linewidth]{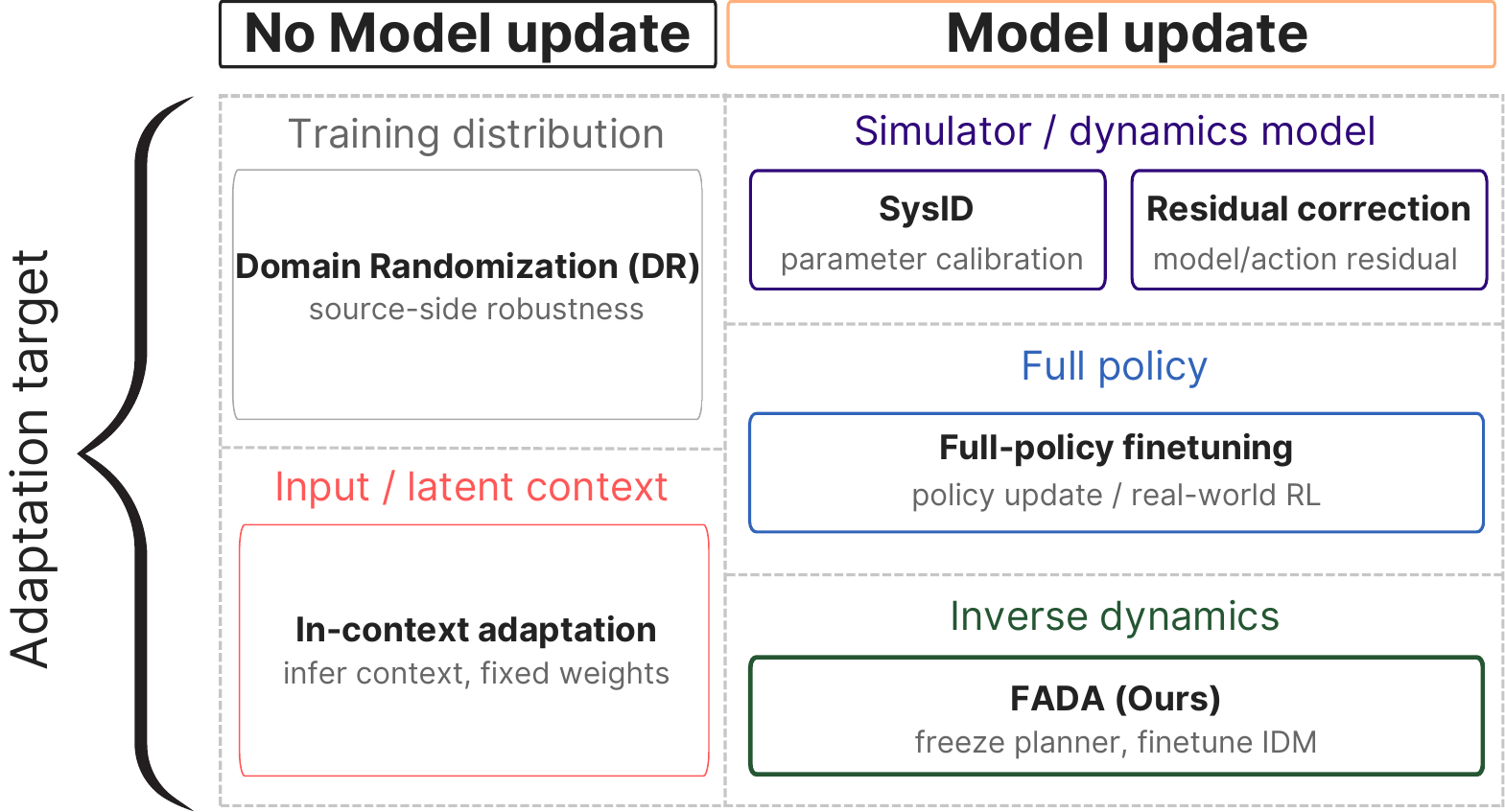}\else\includegraphics[width=\linewidth]{figs/adaptation_taxonomy.pdf}\fi
    \caption{
        \textbf{Adaptation taxonomy.}
        Existing approaches differ in whether they use target rollouts for model updates and which component they update. \idea updates the IDM with target-domain rollouts.}
    \label{fig:adaptation_taxonomy}
    \vspace{-1.0em}
\end{wrapfigure}
Existing approaches for improving target-domain deployment broadly fall into two categories, as summarized in \Cref{fig:adaptation_taxonomy}.
The first category seeks zero-shot robustness or in-context adaptation, including domain randomization~\citep{gu2024humanoidgym,bohlinger2025multiembodiment,cha_sim--real_2025} and history-conditioned adaptation~\citep{kumar_rma_2021,nahrendra_dreamwaq_2023,long_hybrid_2023,liu_locoformer_2025}.
These methods make deployment more robust, but do not update policy weights with target-domain data. They therefore remain conservative under the target condition.
The second category uses target-domain data for adaptation, including system identification~\citep{ren_adaptsim_2023,pmlr-v305-sobanbabu25a,chen_real-time_2022}, residual dynamics learning~\citep{lee_sym2real_2025,he_asap_2025}, and policy finetuning~\citep{smith_legged_2021,hu2025robottrainsrobot,huang_towards_2026}.
These methods can specialize to the target domain, but typically require either a predefined model to fit, new demonstrations to imitate, or a computable target-domain reward for policy optimization. In this work, we ask:

\begin{quote}
   \emph{Is it possible to achieve few-shot target-domain adaptation for precise humanoid control \textbf{without rewards or accurate prior models}?}  
\end{quote}

We answer in the affirmative by introducing \idea: a three-stage Planner--Inverse Dynamics Model (Planner--IDM) framework for few-shot domain adaptation via dynamics alignment.
\idea first trains a privileged oracle policy in simulation, then distills the oracle behavior into a deployable Planner--IDM student through DAgger.
The planner predicts short-horizon proprioceptive intent, and the IDM converts this intent together with recent execution history into actions.
At deployment, \idea collects a small number of target-domain rollouts, freezes the planner, and finetunes only the IDM with supervised learning.
This aligns the IDM with target-domain dynamics while avoiding simulator fitting, real-world reinforcement learning, and full-policy updates.
We evaluate \idea in cross-simulator transfer and real-hardware deployment.
Our experiments cover IsaacSim-to-MuJoCo transfer and deployment on Unitree G1 and Booster T1.
As shown in \Cref{fig:teaser}, after few-shot IDM finetuning, the Planner--IDM controller reliably performs high-precision whole-body tasks on hardware, including locomotion, whole-body tracking, and loco-manipulation.
Results show that \idea improves target-domain performance in both sim-to-sim and sim-to-real settings, demonstrating that supervised IDM finetuning from only a few target-domain rollouts can enable effective adaptation under dynamics shifts.

%% file: sections/related.tex
\section{Related Work}
\label{sec:related_work}

Domain adaptation for humanoid whole-body control is commonly studied as a sim-to-real transfer problem.
As summarized in \Cref{fig:adaptation_taxonomy}, existing methods can be organized by whether they use target-domain data to update the policy, and by which component they adapt: the source training distribution, the inferred deployment-time context, the simulator or dynamics model, the full policy, or the deployed controller.

\subsection{Source-Side Robustness: Domain Randomization and Simulator Alignment}
\label{sec:rw_source_robustness}

A common strategy is to absorb the sim-to-real mismatch before deployment.
\textbf{Domain randomization} trains policies over distributions of physical and sensing parameters, such as mass, friction, actuator response, latency, and observation noise~\cite{gu2024humanoidgym, fu2024humanplus, cheng2024exbody}.
It is now standard in humanoid locomotion and whole-body control~\cite{he2024hover, he2024omnih2o, xie_humanoid_2025, seo_learning_2025}, with recent variants randomizing actuator dynamics, embodiments, or large-scale multi-skill curricula~\cite{cha_sim--real_2025, bohlinger2025multiembodiment, sleiman_zest_2026, wang_omnixtreme_2026}.
However, broad randomization often yields an overly conservative policy rather than one specialized to the deployed system~\cite{da_survey_2025}.
\textbf{Simulator alignment} instead uses target-domain data to align the source simulator or a dynamics model through system identification, residual dynamics, or learned correction models~\cite{ren_adaptsim_2023, pmlr-v305-sobanbabu25a, he_asap_2025, krishna_diffcotune_2025, lee_sym2real_2025}.
These methods provide target-specific corrections, but require simulator fitting and are limited by the chosen parameterization or correction model.
\idea bypasses simulator refitting and adapts the execution module directly from target rollouts while preserving the source policy interface.

\subsection{Inference-Time In-Context Adaptation}
\label{sec:rw_input_adaptation}

A second class of methods adapts at deployment by changing the policy input while keeping policy parameters fixed, corresponding to the ``\emph{input / latent context}'' branch in \Cref{fig:adaptation_taxonomy}.
RMA-style teacher--student methods train a privileged teacher and distill its latent into a deployable history encoder~\cite{kumar_rma_2021}, with extensions to bipedal hardware, humanoid locomotion, dynamic-load adaptation, and uncertainty-aware control~\cite{kumar_adapting_2022, cui_adapting_nodate, chang2025beyond, kumar_error-aware_2021}.
Related world-model approaches replace privileged supervision with self-supervised prediction of terrain, proprioception, future states, or state-action evolution~\cite{nahrendra_dreamwaq_2023, long_hybrid_2023, liu_locoformer_2025, lyu_dywa_2025, zhang_track_2025, xue_lswm_2026, sun_learning_2025, huang_barlowwalk_2025, wang_dadp_2026}.
These methods adapt by updating the inferred context while keeping the action policy fixed.
As a result, they can only exploit target-domain information through a source-trained mapping from context to actions.
In contrast, \idea freezes the planner but finetunes the inverse-dynamics module, allowing target rollouts to directly adapt how planned motions are executed.
\subsection{Target-Domain Learning: Finetuning and Model Adaptation}
\label{sec:rw_target_learning}

A third class of methods uses target-domain rollouts to update the deployed system after transfer.
Residual methods keep a nominal policy fixed and learn an additive action or dynamics correction~\cite{zhao_resmimic_2025, he_asap_2025, cheng_rambo_2025}, but they act at the final action or model level and treat the policy internals as a black box.
Policy finetuning and sim-and-real co-training update the policy itself using target-domain experience, such as real-world RL, human corrections, robot-assisted resets, hybrid online-offline RL, or mixed sim-real training~\cite{smith_legged_2021, jiang_transic_2024, hu2025robottrainsrobot, lei_uni-o4_2024, jones_beyond_2025, maddukuri_sim-and-real_2025, lei_mechanistic_2026}.
These methods can recover performance, but full-policy updates are data-intensive and may entangle command interpretation with low-level execution.
World-model and planning-based methods adapt a learned dynamics model and use it for planning or policy optimization~\cite{levy_simdist_2026, li_robotic_2025, li_uncertainty-aware_2026}, but still require optimization through the learned model at deployment.
\citet{christiano_transfer_2016} execute simulator-predicted next states with an inverse dynamics model learned on the real robot, but require the simulator in the loop at deployment to predict these next states.
In contrast, \idea adapts the execution module inside a factorized policy, requiring no target reward, privileged labels, simulator recalibration, online RL, or MPC-style planning.

%% file: sections/framework_fig.tex
\begin{figure}[t]
    \centering
    \includegraphics[width=\linewidth]{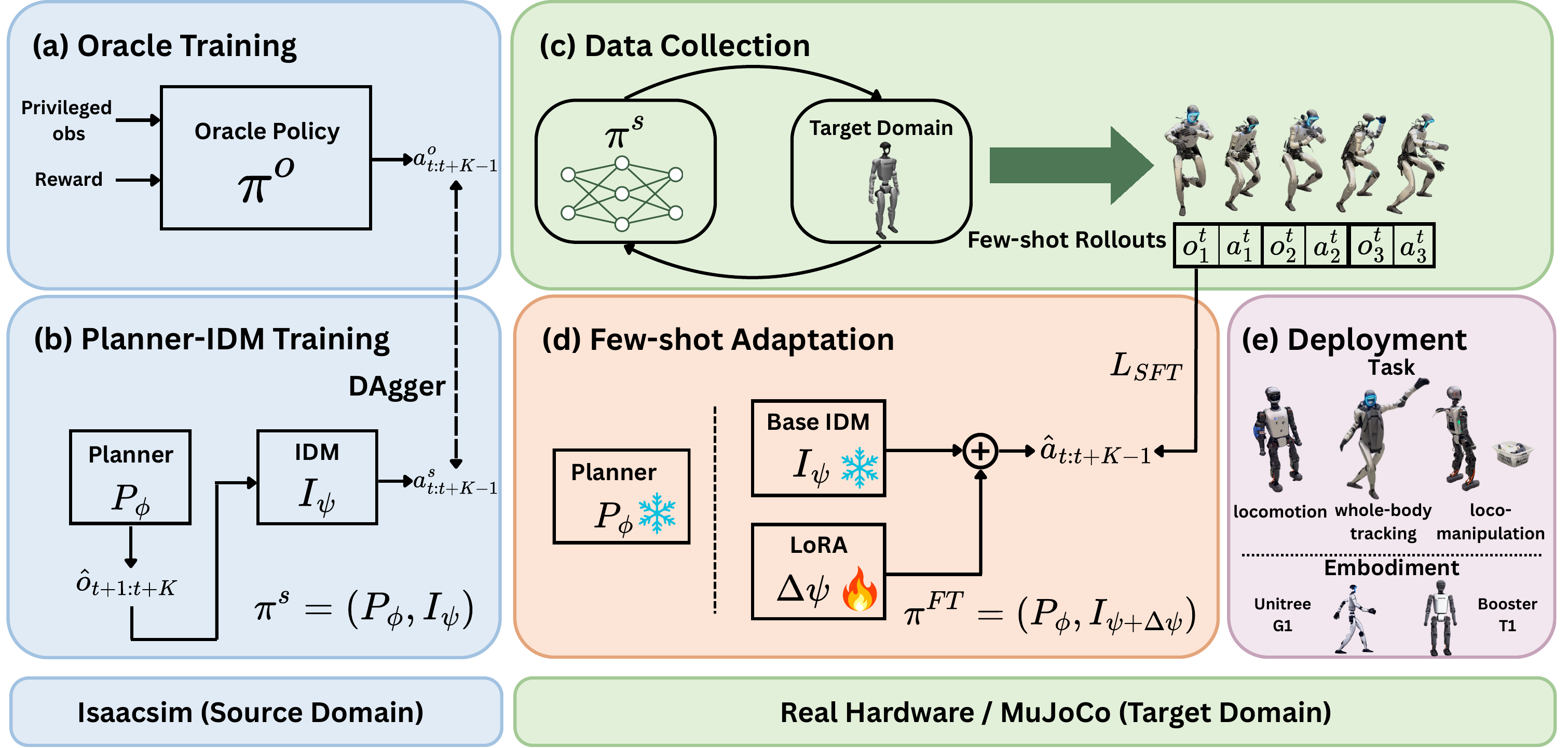}
    \caption{
        \textbf{Overview of \idea.}
        \idea first trains a privileged oracle policy in the source simulator, then distills it into a deployable Planner--IDM student through DAgger-style supervision.
        The planner predicts short-horizon future proprioception from the task command and observation history, while the IDM maps this future to actions.
        During target adaptation, \idea freezes the planner and finetunes only the IDM using a few target-domain rollouts.
        }
    \label{fig:framework_overview}
\end{figure}

%% file: sections/problem_setup.tex
\newcommand{\piorac}{\pi^{\mathrm{o}}}

\section{Problem Setup}
\label{sec:problem}

We study few-shot adaptation for a fixed task under changing deployment dynamics.
Let $\cT$ denote a task, such as whole-body tracking, payload locomotion, or loco-manipulation, and let $\xi$ denote a deployment condition that captures terrain contact, payload, and sensing noise.
For each task and condition, we consider an MDP
$\cM_{\cT,\xi}=(\mathcal{S},\mathcal{O},\mathcal{A},p_{\xi},r_{\cT})$
where $s_t\in\mathcal{S}$ is the full state, $o_t\in\mathcal{O}$ is the proprioceptive observation available at deployment, $a_t\in\mathcal{A}$ is the action, and $c_t\in\mathcal{C}$ denotes the task command (e.g., desired velocity or reference motion). The transition dynamics are given by $p_{\xi}$ and the task objective by $r_{\cT}$.

\paragraph{Source-domain training.}
For each task $\cT$, source training is performed over a distribution of simulated deployment conditions $\xi\sim\srcdom$.
We assume access to a standard privileged-teacher pipeline: a privileged oracle $\pi^{o}(s_t,c_t)\mapsto a_t^{\star}$ is trained in simulation using task rewards and privileged state information, and a student policy is trained from DAgger-style rollouts.
The student receives only proprioceptive observation history, action history, and task commands,
\begin{equation}
\label{eq:student_interface}
    \pi^{s}(\mathcal{O}_t^{H},\mathcal{A}_t^{H},c_t)\mapsto a_t,
    \quad
    \mathcal{O}_t^{H}=(o_{t-H+1},\ldots,o_t),
    \quad
    \mathcal{A}_t^{H}=(a_{t-H},\ldots,a_{t-1}).
\end{equation}
Thus, source training has access to privileged oracle actions as labels and DAgger-style rollouts of proprioceptive observations, commands, and executed actions.

\paragraph{Few-shot target deployment.}
At deployment, the source-trained student is evaluated under a fixed target condition $\xi^{\mathrm{tgt}}$, such as a new simulator, hardware platform, payload, terrain, or contact regime.
The task remains unchanged, but the transition dynamics may differ from those seen during source training.
After zero-shot deployment, we collect a small set of target rollouts
\begin{equation}
\label{eq:target_rollouts}
    \mathcal{D}_{\mathrm{tgt}}
    =
    \{\tau_i\}_{i=1}^{N_{\mathrm{roll}}},
    \qquad
    \tau_i=\{(o_t,c_t,a_t)\}_{t=0}^{T_i}.
\end{equation}
Unlike source training, which may have privileged information, target deployment has access only to these rollouts of proprioceptive observations, commands, and executed actions.
The adaptation problem is to improve the source-trained student under $\xi^{\mathrm{tgt}}$ using only these rollouts, without target-domain rewards, privileged information, expert demonstrations, or simulator fitting.

%% file: sections/method.tex
\section{\idea: Few-Shot Adaptation via Dynamics Alignment}
\label{sec:method}

\begin{wrapfigure}[12]{r}{0.40\linewidth}
    \vspace{-0.6\baselineskip}
    \centering
    \iflatexml\includegraphics[width=0.6\linewidth]{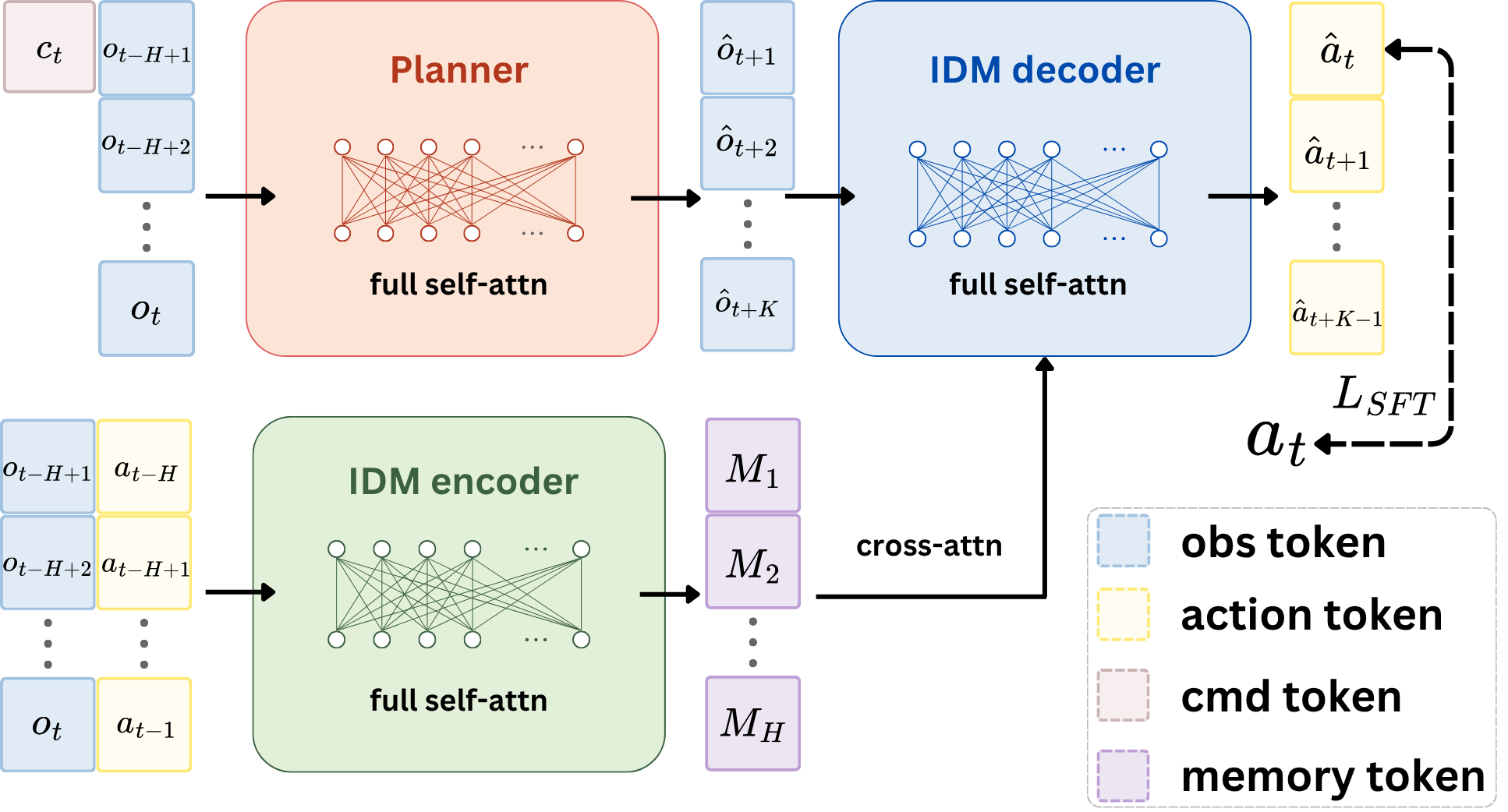}\else\includegraphics[width=\linewidth]{figs/planner_idm_interface.png}\fi
    \caption{\textbf{Planner--IDM interface.} The planner predicts proprioceptive intent, and the IDM maps intent and execution history to an action chunk.}
    \label{fig:planner_idm_arch}
    \vspace{-0.4\baselineskip}
\end{wrapfigure}

Given the source-domain data and few-shot target rollouts defined in \Cref{sec:problem}, \idea is built on a simple observation: under target-domain dynamics shifts, the task intention often remains meaningful, but the action required to realize it can change substantially.
For example, changes in payload or terrain contact may not change what the robot should do, but they can change how the robot must act to produce the intended motion.
This suggests adapting the execution component rather than relearning the entire policy; \Cref{sec:experiments_arm_analysis} validates this separation directly in a controlled setting.
To make this distinction explicit, \idea factorizes the student policy into the Planner--Inverse Dynamics Model (IDM) interface shown in \Cref{fig:planner_idm_arch}: a planner that maps task commands and recent observations to a short-horizon proprioceptive intent, and an IDM that maps this intent and recent execution history to an action chunk.
The overall training, adaptation, and deployment pipeline is summarized in \Cref{fig:framework_overview}.

We instantiate the deployable student policy $\pi^s$ as a Planner--IDM factorization, $\pi^s=(P_\phi,I_\psi)$. At time $t$, the planner $P_\phi$ predicts a $K$-step future proprioceptive sequence
$\hat{Y}_t^K=(\hat{o}_{t+1},\ldots,\hat{o}_{t+K})$, and the IDM $I_\psi$ maps this predicted future, together with recent observation-action history, to an action chunk
$\hat{U}_t^K=(\hat{a}_t,\ldots,\hat{a}_{t+K-1})$.
The factorized student policy is
\begin{equation}
\label{eq:factorized_policy}
    \hat{Y}_{t}^{K}
    =
    P_{\phi}(\mathcal{O}_t^{H}, c_t),
    \qquad
    \hat{U}_t^K
    =
    I_{\psi}(\mathcal{O}_t^{H}, \mathcal{A}_t^{H}, \hat{Y}_{t}^{K}),
    \qquad
    \hat{a}_t
    =
    \Pi_1(\hat{U}_t^K),
\end{equation}
where $\Pi_1$ selects the first action in the chunk.
Although the IDM outputs a full action chunk, deployment follows a receding-horizon interface: only $\hat{a}_t$ is executed before the planner and IDM are queried again.
For source or target rollouts, the same notation applies to realized windows:
$Y_{t,\mathrm{exec}}^K=(o_{t+1},\ldots,o_{t+K})$ and
$U_{t,\mathrm{exec}}^K=(a_t,\ldots,a_{t+K-1})$.
These rollout windows provide inverse-dynamics supervision from the robot's own execution, even when the rollout is not task-optimal.

\subsection{Learning the Planner--IDM Interface}
\label{sec:method_source_interface}

Source training learns two coupled components: an IDM that maps realized proprioceptive futures to the actions executed over the same rollout segment, and a planner that predicts futures that are actionable through the IDM.

\paragraph{IDM learning: realized futures as inverse-dynamics supervision.}
Given a recent observation-action history and the future proprioceptive sequence observed over the same rollout window, the IDM predicts a $K$-step action chunk.
Let $\mathcal{D}_{I}^{\mathrm{src}}$ denote the source rollout windows
$(\mathcal{O}_t^H,\mathcal{A}_t^H,Y_{t,\mathrm{exec}}^K,U_{t,\mathrm{exec}}^K)$ extracted from DAgger-style source rollouts, where $U_{t,\mathrm{exec}}^K$ contains the actions actually executed along the rollout rather than privileged oracle labels.
To match the receding-horizon execution interface, we supervise only the first action of this chunk:
\begin{equation}
\label{eq:idm_loss}
    \mathcal{L}_{I}(\psi)
    =
    \mathbb{E}_{(\mathcal{O}_t^H,\mathcal{A}_t^H,Y_{t,\mathrm{exec}}^K,U_{t,\mathrm{exec}}^K)\sim\mathcal{D}_{I}^{\mathrm{src}}}
    \left[
    \left\|
    \Pi_1\!\left(
    I_\psi(\mathcal{O}_t^H,\mathcal{A}_t^H,Y_{t,\mathrm{exec}}^K)
    \right)
    -
    \Pi_1(U_{t,\mathrm{exec}}^K)
    \right\|_2^2
    \right].
\end{equation}
Here, $\Pi_1(U_{t,\mathrm{exec}}^K)$ is the first action actually executed in the rollout window. It is not necessarily the privileged oracle action.
Thus, the IDM is trained to recover the executed action associated with a realized future, rather than to imitate the oracle directly.
Because the supervision comes from DAgger-style rollouts, the IDM is trained on both teacher-guided and suboptimal student executions.
This makes the same objective applicable to imperfect target rollouts.

\paragraph{Planner learning: actionable rather than observation-regressed futures.}
The planner should produce futures that are useful to the IDM under the same receding-horizon interface used at deployment.
Let $\mathcal{D}_{P}^{\mathrm{src}}$ denote source training samples
$(\mathcal{O}_t^H,\mathcal{A}_t^H,c_t,a_t^\star)$, where $a_t^\star$ is the privileged oracle action for the visited state, obtained by oracle relabeling as described below.
Rather than regressing $P_\phi$ to an oracle future trajectory, we optimize the planner through the current IDM so that the deployed first action matches the privileged oracle action:
\begin{equation}
\label{eq:planner_loss}
    \mathcal{L}_{P}(\phi)
    =
    \mathbb{E}_{(\mathcal{O}_t^H,\mathcal{A}_t^H,c_t,a_t^\star)\sim\mathcal{D}_{P}^{\mathrm{src}}}
    \left[
    \left\|
    \Pi_1\!\left(
    I_{\bar{\psi}}(\mathcal{O}_t^H,\mathcal{A}_t^H,P_\phi(\mathcal{O}_t^H,c_t))
    \right)
    -
    a_t^\star
    \right\|_2^2
    \right],
\end{equation}
where $I_{\bar{\psi}}$ denotes the current IDM with gradients stopped.
This trains the planner to produce action-grounded futures: futures that need not match an oracle observation trajectory, but that produce oracle-consistent actions when passed through the IDM.
After source training, the planner provides a fixed command-to-intent interface, while the IDM provides the execution module that will be adapted in the target domain.

\paragraph{Oracle relabeling on student rollouts.}
For rollouts generated by the intermediate Planner--IDM student, the oracle label $a_t^\star$ is obtained by state relabeling.
At each visited state we save a simulator snapshot; after the rollout, we restore each snapshot and roll the privileged oracle forward for $K$ steps under the same command, producing an oracle action chunk and the corresponding oracle future-observation chunk $(Y_{\mathrm{orac}}^K, U_{\mathrm{orac}}^K)$.
The first action of this chunk supplies the relabeled oracle action $a_t^\star$ used in \Cref{eq:planner_loss}.
To keep IDM supervision causally matched, trajectory-source batches use the realized pair $(Y_{\mathrm{traj}}^K, U_{\mathrm{traj}}^K)$, while oracle-source batches use the oracle-shadow pair $(Y_{\mathrm{orac}}^K, U_{\mathrm{orac}}^K)$.
The oracle-shadow pair is valid inverse-dynamics supervision because it is generated by physically rolling the oracle forward in the shadow simulator, so its future observations and actions form a causally consistent pair under the oracle policy.
Full source-training details are provided in \Cref{app:source_training}.

\paragraph{Architecture.}
Both modules are transformer-based.
The planner is a transformer encoder over history and command tokens that predicts the future chunk as a residual relative to the latest observation.
The IDM is an encoder--decoder in which future-state tokens apply full (non-causal) self-attention over the predicted chunk and cross-attend to the encoded history, and an action head decodes the $K$-step action chunk in parallel.
Architectural details are provided in \Cref{app:architecture}.

\subsection{Few-Shot Target Adaptation}
\label{sec:method_target_adaptation}

At deployment, \idea adapts the IDM using the robot's own target-domain rollouts.
Given $\mathcal{D}_{\mathrm{tgt}}$, we extract supervision windows
\begin{equation}
\label{eq:target_windows}
    \mathcal{W}_{\mathrm{tgt}}
    =
    \left\{
    \left(
    \mathcal{O}_t^{H},
    \mathcal{A}_t^{H},
    Y_{t,\mathrm{exec}}^{K},
    U_{t,\mathrm{exec}}^K
    \right)
    \right\},
\end{equation}
where $Y_{t,\mathrm{exec}}^{K}$ and $U_{t,\mathrm{exec}}^K$ are the future proprioceptive sequence and executed action sequence from the same target rollout segment.
We denote the target-domain update by $\Delta\psi$, which is implemented as LoRA adapters on the IDM.
During adaptation, the planner parameters $\phi$ and the pretrained IDM weights $\psi$ are frozen, and only $\Delta\psi$ is optimized:
\begin{equation}
\label{eq:adapt_loss}
    \ell_{\mathrm{adapt}}(\Delta\psi)
    =
    \mathbb{E}_{(\mathcal{O}_t^{H},\mathcal{A}_t^{H},Y_{t,\mathrm{exec}}^{K},U_{t,\mathrm{exec}}^{K})
    \sim
    \mathcal{W}_{\mathrm{tgt}}}
    \left[
    \left\|
    \Pi_1\!\left(
    I_{\psi+\Delta\psi}(\mathcal{O}_t^{H},\mathcal{A}_t^{H},Y_{t,\mathrm{exec}}^{K})
    \right)
    -
    \Pi_1(U_{t,\mathrm{exec}}^{K})
    \right\|_2^2
    \right].
\end{equation}
The adapted policy is therefore
\begin{equation}
\label{eq:adapted_policy}
    \pi^{\mathrm{ft}} = (P_\phi, I_{\psi+\Delta\psi}).
\end{equation}
At execution, planner-predicted futures $\hat{Y}_t^K$ are passed to the adapted IDM through the same receding-horizon interface in \Cref{eq:factorized_policy}.
Thus, \idea changes only how future task intents are realized under target dynamics, while leaving the planner, command interface, and runtime inputs unchanged.
The complete source-pretraining and target-adaptation procedure is summarized in \Cref{alg:fada_procedure}, and full implementation details are provided in \Cref{app:implementation}.

\begin{nolinenumbers}
\begin{algorithm}[t]
\iflatexml\else\centering\fi
\caption{\textbf{\idea procedure for few-shot domain adaptation.} }
\label{alg:fada_procedure}
\iflatexml
\begin{description}
\item[Input.] Source-domain oracle policy $\pi^{\mathrm{o}}$ trained with privileged observations and task-specific rewards; target domain $\xi_{\mathrm{tgt}}$.
\item[Output.] Target-domain deployable policy $(P_\phi,I_{\psi+\Delta\psi})$.
\end{description}
\textbf{Source pretraining}
\begin{description}
\item[S1:] Rollout the current source policy in the source domain and append visited-state trajectory pairs $(Y_{\mathrm{traj}}^K,U_{\mathrm{traj}}^K)$.
\item[S2:] Query the oracle $\pi^{\mathrm{o}}$ at visited states to attach oracle-shadow pairs $(Y_{\mathrm{orac}}^K,U_{\mathrm{orac}}^K)$ and first-action labels $a_t^\star$.
\item[S3:] Update the IDM $I_\psi$ on the selected matched pair $(Y_I^K,U_I^K)$ using \Cref{eq:idm_loss}.
\item[S4:] Update the planner $P_\phi$ through the fixed IDM using oracle first-action labels and \Cref{eq:planner_loss}.
\item[S5:] Repeat S1--S4 as the DAgger source-pretraining loop.
\end{description}
\textbf{Few-shot target adaptation}
\begin{description}
\item[T1:] Collect target rollouts with the pretrained policy $(P_\phi,I_\psi)$ and extract $\mathcal{W}_{\mathrm{tgt}}$.
\item[T2:] Freeze $P_\phi$ and $\psi$, then optimize only $\Delta\psi$ with \Cref{eq:adapt_loss}.
\item[T3:] Deploy $(P_\phi,I_{\psi+\Delta\psi})$ with planner-predicted futures $\hat{Y}_t^K$ and receding-horizon first-action execution.
\end{description}
\else
\begin{minipage}{0.94\linewidth}
\small
\setlength{\tabcolsep}{2pt}
\renewcommand{\arraystretch}{1.08}
\begin{tabular}{@{}r@{\hspace{0.45em}}p{0.88\linewidth}@{}}
    \multicolumn{2}{@{}p{0.94\linewidth}@{}}{\textbf{Input.} Source-domain oracle policy $\pi^{\mathrm{o}}$ trained with privileged observations and task-specific rewards; target domain $\xi_{\mathrm{tgt}}$.} \\
    \multicolumn{2}{@{}p{0.94\linewidth}@{}}{\textbf{Output.} Target-domain deployable policy $(P_\phi,I_{\psi+\Delta\psi})$.} \\
    \rowcolor{srcband}
    \multicolumn{2}{@{}l@{}}{\textbf{Source pretraining}} \\
    S1: & Rollout the current source policy in the source domain and append visited-state trajectory pairs $(Y_{\mathrm{traj}}^K,U_{\mathrm{traj}}^K)$. \\
    S2: & Query the oracle $\pi^{\mathrm{o}}$ at visited states to attach oracle-shadow pairs $(Y_{\mathrm{orac}}^K,U_{\mathrm{orac}}^K)$ and first-action labels $a_t^\star$. \\
    S3: & Update the IDM $I_\psi$ on the selected matched pair $(Y_I^K,U_I^K)$ using \Cref{eq:idm_loss}. \\
    S4: & Update the planner $P_\phi$ through the fixed IDM using oracle first-action labels and \Cref{eq:planner_loss}. \\
    S5: & Repeat S1--S4 as the DAgger source-pretraining loop. \\
    \rowcolor{realband}
    \multicolumn{2}{@{}l@{}}{\textbf{Few-shot target adaptation}} \\
    T1: & Collect target rollouts with the pretrained policy $(P_\phi,I_\psi)$ and extract $\mathcal{W}_{\mathrm{tgt}}$. \\
    T2: & Freeze $P_\phi$ and $\psi$, then optimize only $\Delta\psi$ with \Cref{eq:adapt_loss}. \\
    T3: & Deploy $(P_\phi,I_{\psi+\Delta\psi})$ with planner-predicted futures $\hat{Y}_t^K$ and receding-horizon first-action execution. \\
\end{tabular}
\end{minipage}
\fi
\end{algorithm}
\end{nolinenumbers}

%% file: sections/experiments.tex
\section{Experiments}
\label{sec:experiments}

Our experiments evaluate \textbf{\idea} as a few-shot adaptation framework for humanoids by addressing the following questions:
 \textbf{(1)} whether few-shot IDM finetuning improves real-world deployment where zero-shot transfer is unreliable, \textbf{(2)} whether the framework improves transfer across embodiments, tasks, and controlled sim-to-sim dynamics shifts relative to in-context and future-prediction adaptation baselines, \textbf{(3)} how large a dynamics shift few-shot IDM adaptation can handle, and whether the adapted IDM transfers to nearby conditions, \textbf{(4)} how the prediction horizon $K$ affects final performance, and which predicted future steps the IDM relies on, and \textbf{(5)} how the planner and IDM training objectives affect whether the zero-shot policy remains deployable enough to collect target rollouts.
We further validate the design choices behind this recipe through adaptation-design ablations and a controlled analysis that directly tests the planner--IDM role separation underlying \idea.

\textbf{Tasks and Systems.}
We evaluate \idea on two humanoid platforms, Unitree~G1 (29~DoF) and Booster~T1 (23~DoF), across sim-to-sim (MuJoCo~\cite{todorov2012mujoco}) and sim-to-real settings.
The task suite covers locomotion and whole-body tracking under dynamics mismatch induced by payload and terrain shifts.
All student policies use history length $H=30$ and prediction horizon $K=6$ and are trained in IsaacSim~\cite{mittal2023orbit} with domain randomization.
Task definitions, deployment conditions, and source-training details are provided in \Cref{app:tasks_conditions,app:source_training}.
For target adaptation, locomotion tasks use approximately two minutes of target rollouts at $50$\,Hz ($\approx 6000$ control steps), and whole-body tracking tasks use six repetitions of the $\approx 20$\,s reference motion, yielding the same budget; the sim-to-sim adaptation budget is matched to this hardware budget.
Unless an ablation states otherwise, sim-to-real results average five hardware trials and main sim-to-sim results average twenty trials.

\begin{figure}[t]
    \centering
    \includegraphics[width=0.98\linewidth]{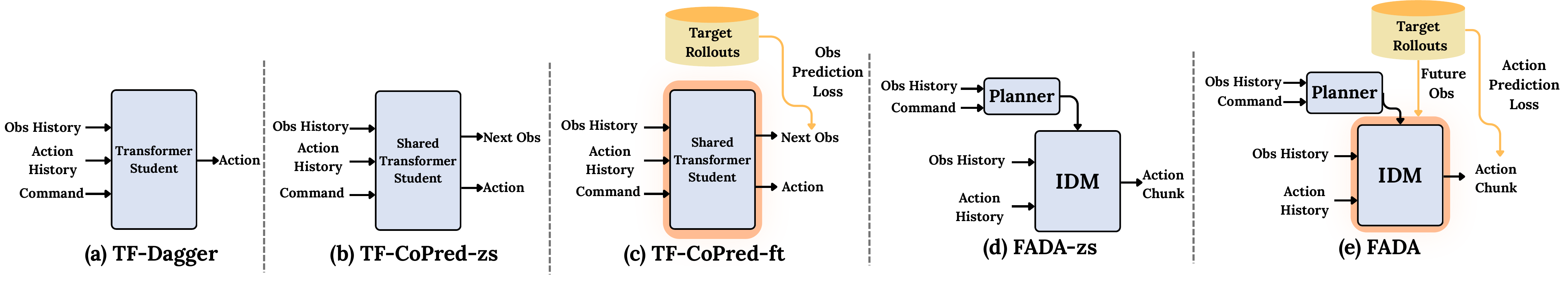}
    \caption{\textbf{Baseline interfaces.} We compare \idea with source-trained transformer DAgger, zero-shot co-prediction, and target-domain co-prediction finetuning. The comparison isolates whether target rollouts are most useful when they update the execution module rather than a monolithic student or a future-prediction objective.}
    \label{fig:baselines}
\end{figure}

\textbf{Baselines and metrics.}
We compare against three teacher--student baselines distilled from the same privileged oracle as \idea, with their interfaces visualized in \Cref{fig:baselines}.
\emph{\tfdagger} is a transformer teacher--student framework~\cite{kumar_rma_2021} that distills a privileged teacher into a history-conditioned policy without target-domain updates.
Motivated by recent world-model and world-action-model approaches that learn dynamics-aware representations through future-state or state-action prediction~\cite{nahrendra_dreamwaq_2023,long_hybrid_2023,lyu_dywa_2025}, \emph{\copred-zs} uses a shared transformer backbone to jointly predict the next observation and action without explicit separation between planning and inverse dynamics.
\emph{\copred-ft} finetunes the shared co-prediction backbone on target rollouts using only a future-observation prediction loss, testing whether end-to-end predictive adaptation can improve deployment without adapting a dedicated action-generation module.
We additionally denote by \emph{\idea-zs} the pre-adaptation Planner--IDM student $(P_\phi, I_\psi)$, i.e., \idea evaluated zero-shot before any target-domain update.
Full baseline implementations are provided in \Cref{app:baselines}.

We report success rate for completion tasks, normalized linear-velocity tracking error $\bar{E}_{v}$ for locomotion, and normalized mean per-joint position error $\bar{E}_{\mathrm{mpjpe}}$ for whole-body tracking.

\subsection{Sim-to-Real Performance after Few-Shot Adaptation}
\label{sec:experiments_sim2real}

\begin{figure}[!t]
    \centering
    
  \includegraphics[width=0.98\linewidth]{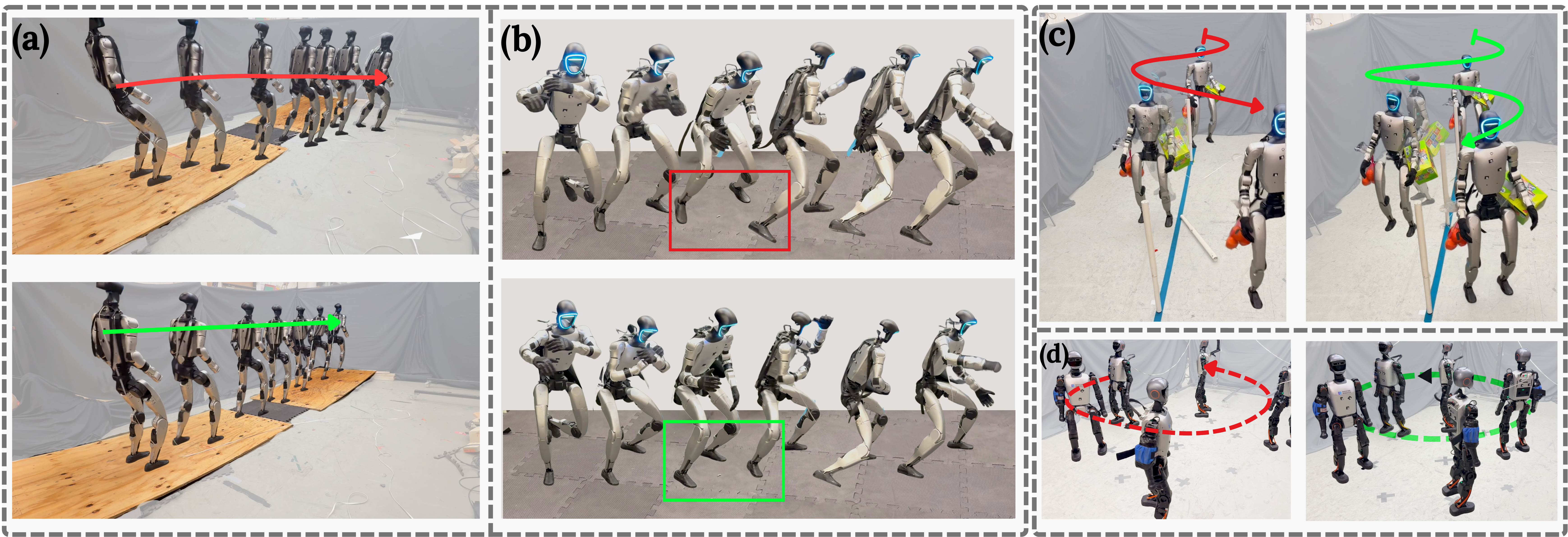}
    \caption{
\textbf{Qualitative sim-to-real deployment.}
Zero-shot and IDM-adapted rollouts for
(a) G1 Slope Traversal, (b) G1 Kungfu + Soft Terrain, (c) G1 Loco. + Payload (grocery carrying through poles), and (d) T1 Loco.\ + Payload.
Adaptation improves execution-critical behavior, including foot placement, posture recovery, and payload compensation.
}   \label{fig:sim2real_rollouts}
\end{figure}

To address \textbf{Q1}, we evaluate \idea on five hardware tasks spanning locomotion and whole-body tracking; \Cref{tab:sim2real_main} reports results and target-domain shifts, with full task definitions in \Cref{app:tasks_conditions}, and \Cref{fig:sim2real_rollouts} shows representative rollouts.
For each task, we collect target rollouts with the pre-adaptation policy \idea-zs, freeze the planner, finetune only the IDM, and re-evaluate over \(5\) hardware trials.
We compare against \idea-zs and the transformer teacher--student baseline \tfdagger.

\input{tables/sim2real_results}

As shown in \Cref{tab:sim2real_main}, few-shot IDM adaptation improves all five hardware tasks.
On success-rate tasks, \idea increases average success from \(20\%\) zero-shot to \(\mathbf{90\%}\), while \tfdagger fails both tasks.
On normalized-error tasks, \idea reduces error by \(\mathbf{27.4\%}\) on average relative to \idea-zs and \(15.9\%\) relative to \tfdagger.
The qualitative rollouts in \Cref{fig:sim2real_rollouts} show that these gains correspond to better execution during slope contacts, payload-induced posture shifts, and soft-terrain tracking.
These performance gains are accompanied by a consistent decrease in IDM loss across all five tasks, indicating that adaptation improves the IDM's alignment with target-domain dynamics.
We compute this loss using the same first-action inverse-dynamics objective as in \Cref{eq:idm_loss}: for each policy, the loss is evaluated on rollout windows collected by that policy in the corresponding target domain.
Thus, lower IDM loss reflects better target-domain action prediction and explains the improved hardware performance.
Notably, on G1 Kungfu + Soft Terrain, \idea improves whole-body tracking despite the deformable mat being outside the source-training distribution.
The zero-shot failures also separate into two recurring modes: under terrain or contact mismatch the policy still initiates the correct skill but accumulates foot-placement and posture errors over time, while under payload or dragging forces the planned motion remains reasonable but the action map can no longer realize the intended body motion.
IDM finetuning addresses both modes by recalibrating the plan-to-action map from target rollouts, and broader qualitative evidence is reported in \Cref{app:sim2real_extra}.
These results show that ordinary target-domain observation-action rollouts are sufficient to improve IDM action prediction under deployment dynamics, without target rewards, privileged labels, simulator recalibration, or full-policy finetuning.

\subsection{Sim-to-Sim Transfer across Embodiments and Tasks}
\label{sec:experiments_sim2sim}

To address \textbf{Q2}, we use IsaacSim-to-MuJoCo transfer as a controlled testbed for studying how target-domain adaptation should be applied.
We evaluate five tasks across G1 and T1, with all adapted methods using the same few-shot MuJoCo rollout budget and only proprioceptive observation-action trajectories.
\input{tables/sim2sim_results}
Two patterns emerge from \Cref{tab:sim2sim_main}.
First, in the \emph{source} IsaacSim$_{\mathrm{src}}$ columns, all zero-shot methods remain close to the \emph{\tfdagger} reference, indicating that the MuJoCo gains are not due to differences in source-domain performance.
Second, under held-out \emph{MuJoCo} dynamics, the advantage appears when target supervision updates the IDM.
Averaged across all five tasks, \idea reduces normalized error by $\mathbf{24.7\%}$ over \idea-zs and $\mathbf{26.8\%}$ over \emph{\tfdagger}.
The largest gain appears on T1 Falcon~\citep{zhang2025falcon}, a force-adaptive locomotion task in which the robot must track commanded velocities while compensating for a persistent external pulling force.
Because performance in this task depends strongly on accurately compensating for external forces, it is especially sensitive to dynamics mismatch and therefore benefits from IDM adaptation.
In contrast, \emph{\copred-ft} worsens relative to \emph{\copred-zs}, suggesting that improving future-observation prediction alone does not guarantee better deployment performance when the action-generation map remains mismatched.
These results support adapting the IDM with target rollouts as a way to improve transfer without changing the planner or policy interface.
These gains are also not an artifact of cross-simulator implementation differences: under same-simulator (IsaacSim-to-IsaacSim) held-out payload and terrain shifts, evaluated over $1024$ parallel environments, \idea still consistently improves over zero-shot transfer (\Cref{app:same_sim_shift}).

\subsection{Adaptation Boundary and Cross-Condition Transfer}
\label{sec:experiments_operating_range}

\input{tables/payload_sweep}
To address \textbf{Q3}, we sweep the torso payload of T1 Loco.\ + Payload from $0$ to $30$\,kg in MuJoCo (\Cref{tab:payload_sweep}), well beyond the source randomization range of $\mathcal{U}(-3,6)$\,kg (\Cref{app:domain_randomization}).
At each payload, \idea is adapted with the same $2$-minute budget; when a rollout falls, its pre-fall segment is kept and collection continues until the budget is reached.
We also evaluate FADA@10\,kg, adapted once at $10$\,kg, at every payload.
\idea substantially improves tracking up to $20$\,kg, even though zero-shot rollouts at $15$--$20$\,kg often terminate early.
Zero-shot task completion is therefore not required: pre-fall segments still provide useful observation-action correspondences.
Recovery becomes partial as the shift grows, and at $30$\,kg the pre-fall segments are too short for reliable adaptation, so the boundary depends on how informative the collected data are, and thus on both the task and the shift.

FADA@10\,kg retains about $90\%$ of the matched improvement at $5$ and $15$\,kg and degrades as the evaluation payload moves farther from $10$\,kg, indicating that the adapted IDM does not overfit to a single payload.

\subsection{Influence of Prediction Horizon on Performance}
\label{sec:experiments_horizon}

The prediction horizon $K$ couples the planner and IDM: the planner predicts $K$ future proprioceptive observations, and the IDM attends to this window to output an action chunk whose first action is deployed.
To address \textbf{Q4}, we study both how much future context is needed and how the IDM uses that context.
We evaluate this through: (i) a sweep over $K \in \{1,6,10,15\}$ on post-adaptation IsaacSim-to-MuJoCo transfer for two representative tasks, G1 Kungfu and T1 Loco.\ + Payload; (ii) a \emph{visible-prefix} study on a trained $K{=}6$ model, where only the first $k$ predicted observations are visible and the remaining $6-k$ are masked, and (iii) a \emph{leave-one-out} study, where each predicted step is masked individually to measure its marginal importance.
\iflatexml
\begin{table}[h]
\centering
\input{tables/horizon_ablation_summary}
\end{table}
\begin{figure}[h]
\centering
\includegraphics[width=0.75\linewidth]{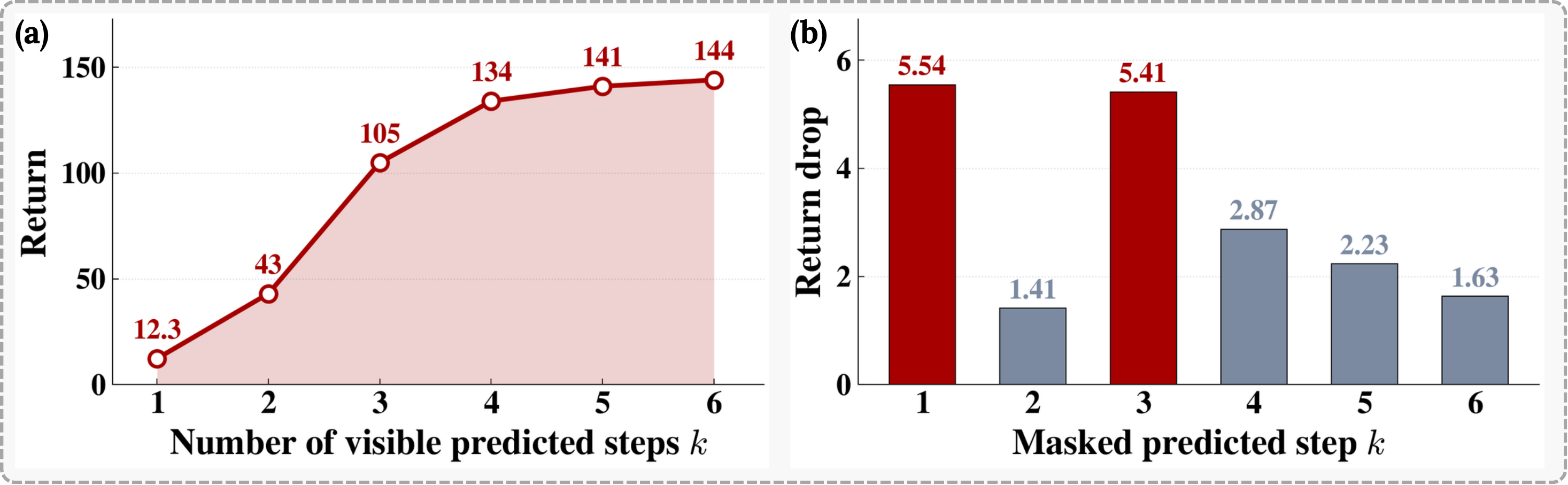}
\caption{\textbf{Attribution over predicted future steps for $\mathbf{K{=}6}$.}
(a) \emph{Visible-prefix}: only the first $k$ predicted observations are given to the IDM.
(b) \emph{Leave-one-out}: one predicted step is masked and the return drop is measured.}
\label{fig:horizon_attribution}
\end{figure}
\else
\begin{center}
    \centering
    \begin{minipage}[t]{0.3\textwidth}
        \vspace{0pt}
        \input{tables/horizon_ablation_summary}
    \end{minipage}\hfill
    \begin{minipage}[t]{0.69\textwidth}
        \vspace{0pt}
        \centering
        \includegraphics[width=\linewidth]{figs/horizon_attribution.pdf}
        \captionof{figure}{
        \textbf{Attribution over predicted future steps for $\mathbf{K{=}6}$.}
        (a) \emph{Visible-prefix}: only the first $k$ predicted observations are given to the IDM.
        (b) \emph{Leave-one-out}: one predicted step is masked and the return drop is measured.
        }
        \label{fig:horizon_attribution}
    \end{minipage}
\end{center}
\fi
\Cref{tab:horizon_ablation} shows that \(K{=}1\) is myopic: moving to \(K{=}6\) reduces normalized error by about \(18\%\) on both tasks, while longer horizons provide no further gain.
This suggests that useful information is concentrated in a compact future window.
\Cref{fig:horizon_attribution} further explains this behavior: The \emph{visible-prefix analysis} shows that most of the gain appears once the IDM sees the first few future tokens, while the \emph{leave-one-out} analysis identifies the \textbf{first} and \textbf{third} predicted steps as the most influential.
Importantly, although the IDM is supervised only on the first executed action (\Cref{sec:method_source_interface}), its decoder uses full attention over the predicted future tokens, allowing later predictions to shape the deployed action.
This suggests that the IDM is not learning a purely one-step inverse model, but a higher-order plan-to-action map that uses short- and medium-horizon predictions to infer local motion trend and execution context.
Consistent with this, replacing the attention-based modules with MLP counterparts leaves source-domain performance nearly unchanged but removes most of the post-adaptation benefit (\Cref{app:design_ablations}), indicating that attention over history and future tokens is what makes the IDM adaptable.

\subsection{Effect of Planner and IDM Training Objectives on Zero-Shot Transfer}
\label{sec:experiments_planner_loss}

\begin{wrapfigure}[14]{r}{0.40\linewidth}
    \vspace{-0.3\baselineskip}
    \centering
    \iflatexml\includegraphics[width=0.6\linewidth]{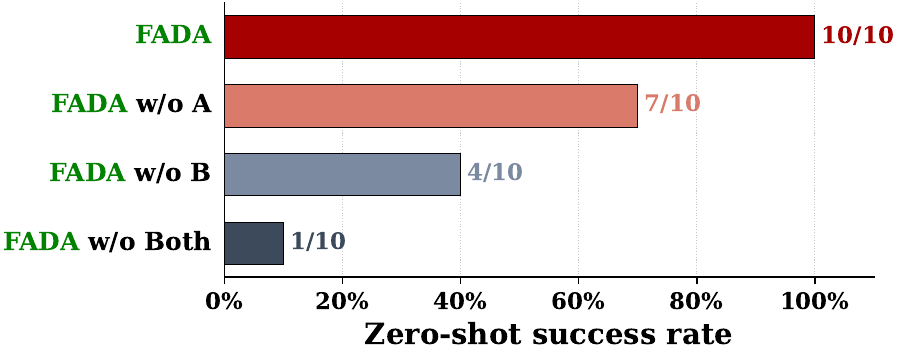}\else\includegraphics[width=\linewidth]{figs/planner_idm_design_ablation.pdf}\fi
    \caption{\textbf{Zero-shot transfer to MuJoCo on G1 whole-body tracking ($n=10$).}
    Our loss formulation in \Cref{sec:method} has two design choices: (A) training the planner through the stop-gradient IDM via action-prediction loss (Eq.~(4.3)); (B) supervising the IDM only on the executed first action (Eq.~(4.2)).
    }
    \label{fig:design_ablation}
    \vspace{-0.3\baselineskip}
\end{wrapfigure}

Few-shot adaptation requires the zero-shot policy to remain deployable long enough to collect target rollouts.
We therefore ablate on two key design choices in our training losses from \Cref{sec:method}, \Cref{eq:idm_loss,eq:planner_loss} that affect zero-shot transfer before any target update is applied.
(A) trains the planner through a frozen IDM using action-prediction loss, rather than directly regressing the planner to oracle future observations (B) supervises the IDM only on the first executed action, rather than on the full predicted action window.
We evaluate \idea and the three ablations on zero-shot IsaacSim-to-MuJoCo transfer for G1 whole-body tracking.
As shown in \Cref{fig:design_ablation}, removing (B) causes the larger drop, reducing success from $\mathbf{10/10}$ to $\mathbf{4/10}$, indicating that full-window action supervision can hurt receding-horizon deployment when only the first action is executed.
Removing (A) reduces success to $7/10$, suggesting that observation-regressed planner outputs can be plausible but not necessarily actionable by the IDM.
Removing both choices compounds the failure to $\mathbf{1/10}$.
Together, these results show that (A) keeps planner predictions compatible with the IDM, while (B) aligns IDM training with the action actually applied by the robot.

\subsection{Adaptation Design Ablations}
\label{sec:experiments_adaptation_design}

We next ablate the two design choices that define \idea's adaptation recipe: updating the IDM through low-rank adapters, and the size of the target rollout budget.
Both ablations are run on MuJoCo sim-to-sim targets with $20$ randomized-command trials per task, and results are normalized column-wise by the first row.

\iflatexml
\begin{table}[h]
\centering
\caption{\textbf{LoRA vs. full IDM finetuning.} We report $\bar E_v\downarrow$ on three MuJoCo target tasks. LoRA provides the most stable few-shot adaptation, while full IDM finetuning can overfit and degrade the policy.}
\label{tab:lora_full_finetune}
\footnotesize
\begin{tabular}{@{}lccc@{}}
        \toprule
        Method & \makecell{G1 Slope\\Traversal} & \makecell{T1 Loco.\\+ Payload} & \makecell{T1 Slope\\Traversal} \\
        \midrule
        \idea-zs & 1.0000$_{\pm0.3788}$ & 1.0000$_{\pm0.1413}$ & 1.0000$_{\pm0.3230}$ \\
        Full IDM FT & 1.1000$_{\pm0.4484}$ & 1.1163$_{\pm0.3342}$ & 1.2918$_{\pm0.2819}$ \\
        LoRA IDM (ours) & \goodnumber{0.8463$_{\pm0.2493}$} & \goodnumber{0.8489$_{\pm0.1293}$} & \goodnumber{0.8846$_{\pm0.1870}$} \\
        \bottomrule
        \end{tabular}
\end{table}
\begin{figure}[h]
\centering
\includegraphics[width=0.6\linewidth]{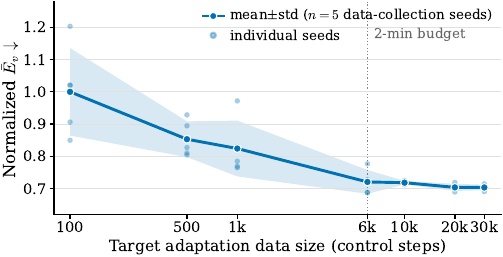}
\caption{\textbf{Target data-size ablation.} We report $\bar E_v\downarrow$ on T1 Loco.\ + Payload, normalized by the $100$-step setting; mean $\pm$ std over five independent data-collection seeds. The $6000$-step budget used in the main experiments reaches the performance plateau, and larger budgets show diminishing returns.}
\label{fig:data_size}
\end{figure}
\else
\begin{center}
    \centering
    \begin{minipage}[t]{0.48\textwidth}
        \vspace{0pt}
        \centering
        \captionof{table}{\textbf{LoRA vs. full IDM finetuning.} We report $\bar E_v\downarrow$ on three MuJoCo target tasks. LoRA provides the most stable few-shot adaptation, while full IDM finetuning can overfit and degrade the policy.}
        \label{tab:lora_full_finetune}
        \footnotesize
        \setlength{\tabcolsep}{2.2pt}
        \renewcommand{\arraystretch}{1.02}
        \resizebox{\linewidth}{!}{%
        \begin{tabular}{@{}lccc@{}}
        \toprule
        Method & \makecell{G1 Slope\\Traversal} & \makecell{T1 Loco.\\+ Payload} & \makecell{T1 Slope\\Traversal} \\
        \midrule
        \idea-zs & 1.0000$_{\pm0.3788}$ & 1.0000$_{\pm0.1413}$ & 1.0000$_{\pm0.3230}$ \\
        Full IDM FT & 1.1000$_{\pm0.4484}$ & 1.1163$_{\pm0.3342}$ & 1.2918$_{\pm0.2819}$ \\
        LoRA IDM (ours) & \goodnumber{0.8463$_{\pm0.2493}$} & \goodnumber{0.8489$_{\pm0.1293}$} & \goodnumber{0.8846$_{\pm0.1870}$} \\
        \bottomrule
        \end{tabular}%
        }
    \end{minipage}\hfill
    \begin{minipage}[t]{0.48\textwidth}
        \vspace{0pt}
        \centering
        \includegraphics[width=\linewidth]{figs/target_rollout_budget_ablation_multiseed.pdf}
        \captionof{figure}{\textbf{Target data-size ablation.} We report $\bar E_v\downarrow$ on T1 Loco.\ + Payload, normalized by the $100$-step setting; mean $\pm$ std over five independent data-collection seeds. The $6000$-step budget used in the main experiments reaches the performance plateau, and larger budgets show diminishing returns.}
        \label{fig:data_size}
    \end{minipage}
\end{center}
\fi

\paragraph{LoRA vs. full IDM finetuning.}
\Cref{tab:lora_full_finetune} compares the default LoRA adaptation against directly finetuning all IDM parameters.
LoRA is consistently the strongest variant across all three target tasks.
Full IDM finetuning, in contrast, often degrades performance relative to the zero-shot policy, indicating that updating the entire action-generation module overfits to the limited target rollout budget.
This is consistent with \idea's premise that the few-shot regime requires a constrained execution-side update rather than a full retraining of the IDM.

\paragraph{Target rollout budget.}
\Cref{fig:data_size} varies the amount of target rollout data on T1 Loco.\ + Payload.
Performance improves quickly from very small budgets to the $6000$-step setting used in the main experiments.
Beyond this point, the mean error decreases only slightly and its variance across five independent data-collection seeds shrinks.
This supports the few-shot framing of \idea: roughly two minutes of target rollouts capture most of the adaptation benefit, and collecting substantially more data brings only marginal additional gain.

\subsection{Controlled Analysis of Planner--IDM Role Separation}
\label{sec:experiments_arm_analysis}

\begin{wrapfigure}[17]{r}{0.36\linewidth}
    \vspace{-0.75\baselineskip}
    \centering
    \iflatexml\includegraphics[width=0.36\linewidth]{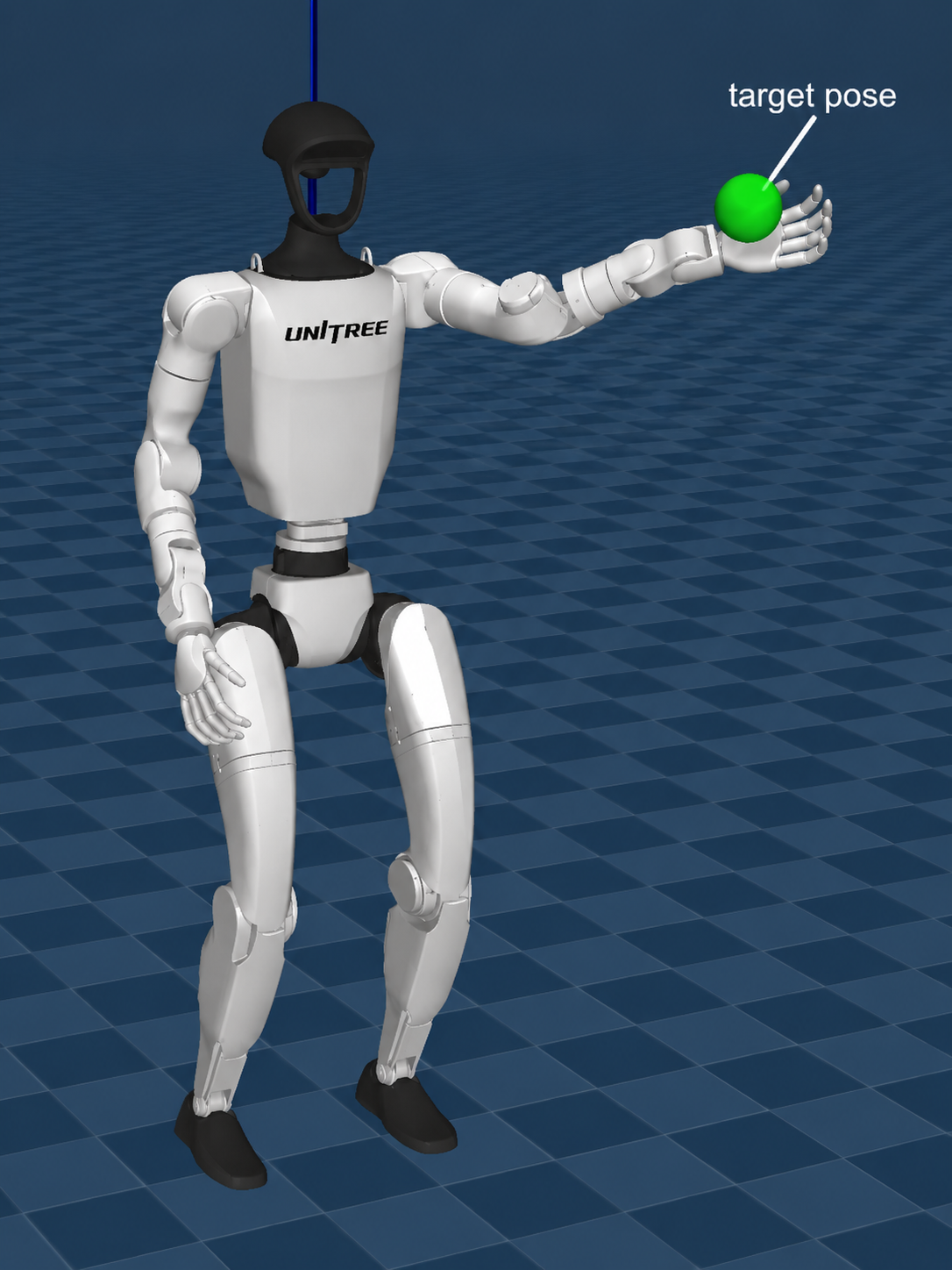}\else\includegraphics[width=\linewidth]{figs/arm_tracking_task.png}\fi
    \caption{\textbf{Arm-tracking task.}
    A fixed-base arm tracks end-effector targets under wrist payloads.}
    \label{fig:arm_task_vis}
    \vspace{-0.65\baselineskip}
\end{wrapfigure}

Finally, we use a fixed-base arm task as a controlled diagnostic for the central hypothesis of \Cref{sec:method}: that dynamics shifts change how an intent must be executed, not the intent itself.
A 7-DoF arm tracks a sequence of Cartesian end-effector targets while the only domain shift is a wrist payload $m \in \{0,\,2.5,\,5.0\}$\,kg (\Cref{fig:arm_task_vis}).
The source model is trained with payload randomization over $m \in [0,\,1.5]$\,kg, so the heavier payloads test extrapolation beyond the source distribution.
This setting separates kinematics from dynamics: the joint configuration needed to reach a target is unchanged by the payload, whereas the action required to realize and hold that configuration changes with mass.
The payload-induced gravity torque is configuration-dependent, so successful adaptation requires more than a constant action offset.

\setlength{\columnsep}{0.55em}

We evaluate three diagnostics before and after few-shot LoRA finetuning of the IDM, with the planner frozen: end-effector tracking error, planner prediction RMSE, and an IDM consistency gap.
The consistency gap compares the IDM action produced using the planner-predicted next observation with the IDM action produced using the actual next observation from the same rollout.
A smaller gap indicates that the executed trajectory is closer to the planner-predicted trajectory, making the deployed and teacher-forced IDM inputs more consistent.

\begin{figure*}[t]
    \centering
    \includegraphics[width=0.92\linewidth]{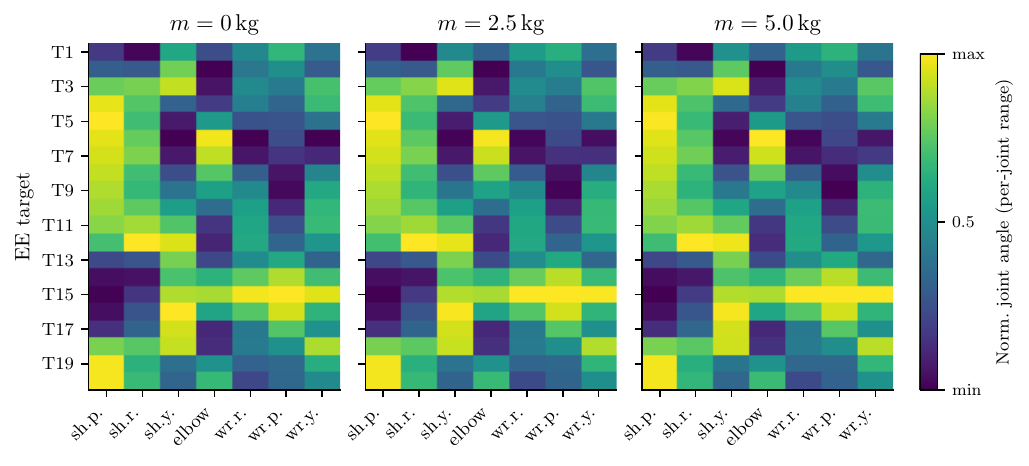}
    \caption{\textbf{Planner IK map across payload conditions before finetuning.}
    Each cell shows the mean joint angle predicted by the planner for a target and joint, with colors normalized per joint across all payloads.
    The near-identical patterns show that the planner maps the same end-effector target to the same joint configuration regardless of payload.}
    \label{fig:arm_ik_heatmap}
\end{figure*}

\paragraph{Planner invariance.}
\Cref{fig:arm_ik_heatmap} shows that the planner predicts nearly identical joint configurations for the same end-effector targets across all payloads.
The mean per-joint planner RMSE changes by only \goodnumber{7\%} from $m{=}0$ to $m{=}5$\,kg ($0.049$ vs.\ $0.052$\,rad), despite the heavier payload being outside the training range.
This supports the intended role of the planner: it captures a payload-invariant kinematic map rather than compensating for payload-dependent dynamics.

\begin{figure*}[t]
    \centering
    \includegraphics[width=\linewidth]{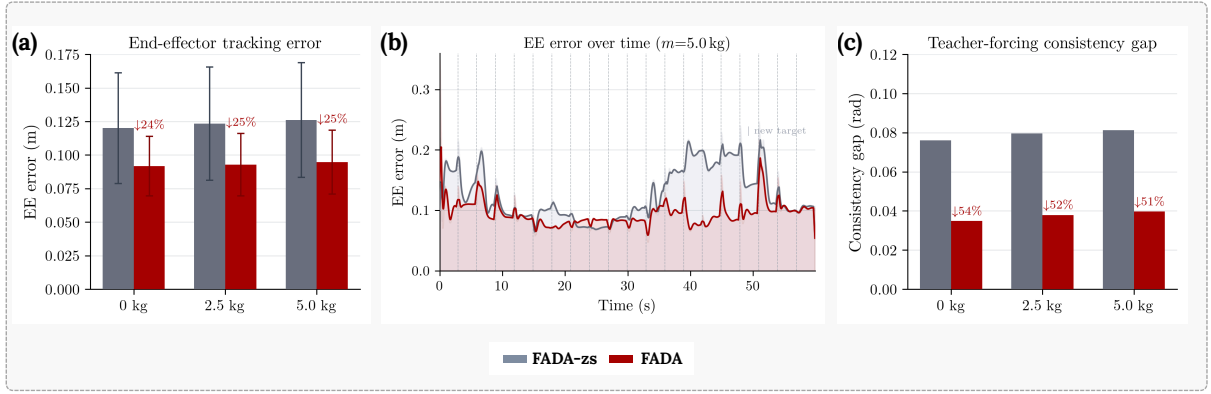}
    \caption{\textbf{Fixed-base arm tracking under payload variation.}
    \textbf{(a)} End-effector tracking error before and after IDM finetuning.
    \textbf{(b)} Per-timestep EE error at $m{=}5.0$\,kg; vertical dashes mark command switches every $3$\,s.
    \textbf{(c)} IDM consistency gap between actions computed from planner-predicted and actual next observations.
    Annotated percentages indicate reductions after adaptation.}
    \label{fig:arm_tracking_main}
\end{figure*}

\paragraph{IDM adaptation.}
\Cref{fig:arm_tracking_main} shows that finetuning only the IDM improves both tracking and interface consistency.
Across payloads, the consistency gap decreases by \goodnumber{$\approx$54\%} on average, from $0.079$ to $0.037$\,rad, while end-effector tracking error decreases by \goodnumber{$\approx$24\%}.
The planner RMSE also decreases by \goodnumber{$\approx$39\%}, even though the planner weights are frozen.
This does not indicate that the planner changed; rather, the adapted IDM drives the arm toward states that better match the planner's original predictions.
The improvement is not merely a payload-specific action bias, because the wrist load induces configuration-dependent gravity torques, so the required correction varies with the target, posture, and transient motion between targets.
The reduction in both tracking error and consistency gap therefore suggests that IDM finetuning learns a structured, plan-conditioned inverse-dynamics correction using the predicted future and recent execution history.
While this separation is cleanest in the fixed-base setting, it is consistent with the humanoid results in \Cref{sec:experiments_sim2real,sec:experiments_sim2sim}, where adaptation gains concentrate in execution-critical behaviors such as foot placement, posture recovery, and payload compensation.
Overall, this controlled task supports the role separation used by \idea: the planner retains the command-to-kinematics map, while few-shot IDM adaptation recalibrates the dynamics-dependent plan-to-action map.

%% file: tables/sim2real_results.tex
\begin{table}[H]
\caption{Comparison of normalized sim-to-real task performance across methods. For normalized tracking metrics, \mbox{\tfdagger} corresponds to $1.0$ and lower is better. Success is defined as completing the out-and-back path without leaving the ramp (Slope Traversal) and pulling the basket across the finish line (Basket Pulling); full task definitions are given in \Cref{app:tasks_conditions}.}
\label{tab:sim2real_main}
\centering
\resizebox{\linewidth}{!}{%
\begingroup
\setlength{\tabcolsep}{3pt}
\renewcommand{\arraystretch}{0.95}
\begin{tabular}{lcccccccccc}
\toprule
\multicolumn{1}{c}{\textbf{Method}}
& \multicolumn{2}{c}{\makecell{G1 Slope Traversal\\\scriptsize two $15^\circ$ ramps, $\approx0.8$\,m wide}}
& \multicolumn{2}{c}{\makecell{G1 Loco. + Payload\\\scriptsize $\approx3$\,kg asymmetric grocery payload}}
& \multicolumn{2}{c}{\makecell{G1 Kungfu + Soft Terrain\\\scriptsize deformable soft mats}}
& \multicolumn{2}{c}{\makecell{T1 Loco. + Payload\\\scriptsize $1$\,kg asymmetric arm payload}}
& \multicolumn{2}{c}{\makecell{T1 Basket Pulling\\\scriptsize $\approx6$\,kg loaded laundry basket}} \\
\cmidrule(lr){2-3}
\cmidrule(lr){4-5}
\cmidrule(lr){6-7}
\cmidrule(lr){8-9}
\cmidrule(lr){10-11}
& \makecell{IDM loss\\$(\times 1\mathrm{e}{-3})\downarrow$}
& \makecell{Success\\$\uparrow$}
& \makecell{IDM loss\\$(\times 1\mathrm{e}{-3})\downarrow$}
& \makecell{$\bar{E}_{v}\downarrow$}
& \makecell{IDM loss\\$(\times 1\mathrm{e}{-3})\downarrow$}
& \makecell{$\bar{E}_{\mathrm{mpjpe}}\downarrow$}
& \makecell{IDM loss\\$(\times 1\mathrm{e}{-3})\downarrow$}
& \makecell{$\bar{E}_{v}\downarrow$}
& \makecell{IDM loss\\$(\times 1\mathrm{e}{-3})\downarrow$}
& \makecell{Success\\$\uparrow$} \\
\midrule
\emph{\tfdagger}
& -- 
& $0\%$ 
& -- 
& $1.000_{\pm 0.203}$ 
& -- 
& $1.00_{\pm 0.091}$ 
& -- 
& $1.000_{\pm 0.132}$ 
& -- 
& $0\%$ \\

\idea-zs
& $2.391_{\pm 1.877}$
& $20\%$ 
& $2.195_{\pm 0.465}$
& $1.289_{\pm 0.711}$ 
& $5.247_{\pm 0.166}$
& $1.18_{\pm 0.103}$ 
& $1.068_{\pm 0.235}$
& $1.043_{\pm 0.025}$ 
& $0.873_{\pm 0.171}$
& $20\%$ \\

\idea
& \goodnumber{$2.190_{\pm 1.416}$}
& \goodnumber{$80\%$} 
& \goodnumber{$1.848_{\pm 0.282}$}
& \goodnumber{$0.797_{\pm 0.018}$} 
& \goodnumber{$4.460_{\pm 0.141}$}
& \goodnumber{$0.86_{\pm 0.084}$} 
& \goodnumber{$0.954_{\pm 0.197}$}
& \goodnumber{$0.866_{\pm 0.040}$} 
& \goodnumber{$0.650_{\pm 0.0923}$}
& \goodnumber{$100\%$} \\
\bottomrule
\end{tabular}
\endgroup
}
\end{table}

%% file: tables/sim2sim_results.tex
\begin{table*}[t]
\caption{\textbf{Sim-to-sim transfer across embodiments and tasks.}
For each task and method we report normalized task error (lower is better) under two evaluation conditions: the source IsaacSim training condition (\colorbox{srcband}{\strut IsaacSim$_{\mathrm{src}}$}) and a held-out MuJoCo simulator (\colorbox{realband}{\strut MuJoCo}).
Pre-adaptation rows use only source training data; finetuned rows update with the same target-domain rollout budget on a few-shot deployment trajectory.
}
\label{tab:sim2sim_main}
\centering
\begingroup
\small
\resizebox{\linewidth}{!}{%
\setlength{\tabcolsep}{2.8pt}
\renewcommand{\arraystretch}{0.95}
\begin{tabular}{l cc cc cc cc cc}
\toprule
\multicolumn{1}{c}{Task} & \multicolumn{2}{c}{G1 Slope Traversal $\downarrow$} & \multicolumn{2}{c}{G1 Kungfu $\downarrow$} & \multicolumn{2}{c}{T1 Loco. + Payload $\downarrow$} & \multicolumn{2}{c}{T1 Slope Traversal $\downarrow$} & \multicolumn{2}{c}{T1 Falcon $\downarrow$} \\
\cmidrule(lr){2-3} \cmidrule(lr){4-5} \cmidrule(lr){6-7} \cmidrule(lr){8-9} \cmidrule(lr){10-11}
Method &
\cellcolor{srcband}IsaacSim$_{\mathrm{src}}$ & \cellcolor{realband}MuJoCo &
\cellcolor{srcband}IsaacSim$_{\mathrm{src}}$ & \cellcolor{realband}MuJoCo &
\cellcolor{srcband}IsaacSim$_{\mathrm{src}}$ & \cellcolor{realband}MuJoCo &
\cellcolor{srcband}IsaacSim$_{\mathrm{src}}$ & \cellcolor{realband}MuJoCo &
\cellcolor{srcband}IsaacSim$_{\mathrm{src}}$ & \cellcolor{realband}MuJoCo \\
\midrule
\emph{\tfdagger}
& 1.000$_{\pm 0.390}$ & 1.000$_{\pm 0.439}$
& 1.000$_{\pm 0.049}$ & 1.000$_{\pm 0.052}$
& 1.000$_{\pm 0.221}$ & 1.000$_{\pm 0.143}$
& 1.000$_{\pm 0.261}$ & 1.000$_{\pm 0.252}$
& 1.000$_{\pm 0.259}$ & 1.000$_{\pm 0.086}$ \\

\emph{\copred-zs}
& 1.036$_{\pm 0.383}$ & 0.973$_{\pm 0.387}$
& 1.024$_{\pm 0.073}$ & 1.031$_{\pm 0.068}$
& 0.991$_{\pm 0.219}$ & 1.029$_{\pm 0.142}$
& 1.026$_{\pm 0.287}$ & 1.080$_{\pm 0.224}$
& 0.974$_{\pm 0.251}$ & 0.943$_{\pm 0.133}$ \\

\emph{\copred-ft}
& -- & 1.611$_{\pm 0.488}$
& -- & 1.421$_{\pm 0.198}$
& -- & 1.737$_{\pm 0.387}$
& -- & 1.193$_{\pm 0.246}$
& -- & 1.054$_{\pm 0.162}$ \\

\idea-zs
& 1.038$_{\pm 0.386}$ & 0.946$_{\pm 0.358}$
& 0.988$_{\pm 0.050}$ & 0.961$_{\pm 0.055}$
& 1.064$_{\pm 0.208}$ & 1.042$_{\pm 0.147}$
& 1.019$_{\pm 0.243}$ & 1.034$_{\pm 0.334}$
& 0.994$_{\pm 0.316}$ & 0.880$_{\pm 0.094}$ \\

\idea
& -- & \goodnumber{0.800$_{\pm 0.236}$}
& -- & \goodnumber{0.714$_{\pm 0.048}$}
& -- & \goodnumber{0.885$_{\pm 0.135}$}
& -- & \goodnumber{0.914$_{\pm 0.193}$}
& -- & \goodnumber{0.347$_{\pm 0.043}$} \\
\bottomrule
\end{tabular}
}
\endgroup
\end{table*}

%% file: tables/payload_sweep.tex
\begin{table}[t]
\caption{\textbf{Payload sweep on T1 Loco.\ + Payload (MuJoCo).} Cells: normalized $\bar{E}_v\downarrow$ / survival rate (fraction of commanded steps before failure). \idea adapts at the evaluated payload; FADA@10\,kg adapts at $10$\,kg.}
\label{tab:payload_sweep}
\centering
\small
\setlength{\tabcolsep}{6pt}
\renewcommand{\arraystretch}{1.05}
\begin{tabular}{@{}lcccccc@{}}
\toprule
Payload (kg) & 0 & 5 & 10 & 15 & 20 & 30 \\
\midrule
\idea-zs & 1.00\,/\,1 & 1.20\,/\,.95 & 1.31\,/\,.94 & 1.98\,/\,.62 & 2.44\,/\,.43 & 2.56\,/\,.12 \\
\idea & \goodnumber{.85}\,/\,1 & \goodnumber{.87}\,/\,1 & \goodnumber{.93}\,/\,1 & \goodnumber{.99}\,/\,1 & \goodnumber{1.28}\,/\,.94 & \goodnumber{2.40}\,/\,.36 \\
FADA@10\,kg & .88\,/\,1 & .90\,/\,1 & .93\,/\,1 & 1.09\,/\,.98 & 1.60\,/\,.80 & 2.53\,/\,.19 \\
\bottomrule
\end{tabular}
\end{table}

%% file: tables/horizon_ablation_summary.tex
\vspace{1pt}
\centering
{\scriptsize
\begingroup
\setlength{\tabcolsep}{3pt}
\renewcommand{\arraystretch}{1.05}
\begin{tabular}{c cc}
\toprule
$K$ & \cellcolor{realband}\shortstack{G1 WBT\\($\bar{E}_{\mathrm{mpjpe}}\downarrow$)} & \cellcolor{realband}\shortstack{T1 Locomotion\\($\bar{E}_{v}\downarrow$)} \\
\midrule
1  & 1.000$\pm$0.136 & 1.000$\pm$0.129 \\
6  & \goodnumber{0.830$\pm$0.107} & \goodnumber{0.813$\pm$0.122} \\
10 & 0.852$\pm$0.101 & 0.931$\pm$0.121 \\
15 & 0.858$\pm$0.135 & 0.926$\pm$0.112 \\
\bottomrule
\end{tabular}
\endgroup
\par}
\captionof{table}{\textbf{Horizon $K$ ablation (post-adaptation).} Normalized error (lower is better) for G1 tracking ($\bar{E}_{\mathrm{mpjpe}}$) and T1 Loco.\ + Payload ($\bar{E}_{v}$) after few-shot adaptation.}
\label{tab:horizon_ablation}

%% file: sections/conclusion_and_limitations.tex
\section{Conclusion}
\label{sec:conclusion}

We presented \idea, a few-shot domain adaptation framework for humanoid control that aligns target-domain dynamics using only deployment rollouts. \idea factorizes a deployable policy into a planner that specifies future task intent and an IDM that realizes this intent as actions, then adapts only the IDM to the target domain. In both controlled sim-to-sim transfer and real-world deployment on Unitree G1 and Booster T1, \idea improves task performance under dynamics shifts and enables real humanoid robots to execute diverse high-precision whole-body tasks. These results suggest that many humanoid transfer failures are not failures of task intent, but failures of realizing that intent under new physics. By separating intent generation from execution, \idea offers a practical path toward scalable post-deployment adaptation for high-precision humanoid control.

\section{Limitations}
\label{sec:limitations}

While these results are encouraging, the current formulation also has several important limitations.

\paragraph{Dependence on non-trivial zero-shot performance.}
\idea requires the source-trained policy to collect target rollouts with useful observation-action correspondences.
Catastrophic zero-shot failures may therefore require safety mechanisms, recovery controllers, or assisted data collection.

\paragraph{Task-dependent execution module.}
Although the IDM is intended to capture execution dynamics, it is trained within a particular task distribution and may still encode task-specific structure.
A more task-agnostic execution model that transfers across tasks could further reduce adaptation cost and broaden the applicability of the framework.

\paragraph{Proprioception-only adaptation.}
The current instantiation adapts from proprioceptive observations and executed actions without explicitly representing external factors such as terrain geometry or payload distribution. While this avoids environment reconstruction, extending the same principle to richer perceptual inputs or learned world-model representations could improve adaptation when the mismatch is primarily exteroceptive.

%% file: sections/acknowledgements.tex
\section*{Acknowledgements}
\addcontentsline{toc}{section}{Acknowledgements}

We thank Yuxiang Yang for his insightful feedback and suggestions.

%% file: sections/appendix.tex
\section*{Appendix}
\addcontentsline{toc}{section}{Appendix}
\raggedbottom

\section{Experimental Setup}
\label[appendix]{app:experimental_setup}

This appendix provides additional setup details for the experiments in \Cref{sec:experiments}.
We focus on the task definitions, deployment shifts, data budgets, and metrics needed to reproduce the comparisons in the main text.
All source policies are trained in IsaacSim, while target evaluation is performed either on hardware or in held-out simulator conditions.

\subsection{Tasks and Deployment Conditions}
\label[appendix]{app:tasks_conditions}

Each experiment is defined by a task $\cT$, a source condition
$\xi_{\mathrm{src}} \sim \srcdom$, and a target deployment condition
$\xi_{\mathrm{tgt}}$. The task objective remains unchanged across domains, while the transition
dynamics may differ because of changes in payload, terrain contact,
actuator behavior, latency, or simulator implementation.
The goal is to adapt a policy trained under $\srcdom$ so that it
maintains task performance under $\xi_{\mathrm{tgt}}$ using only a small
amount of target-domain interaction data.

\paragraph{Hardware deployment tasks.}
For sim-to-real evaluation, we deploy on Unitree~G1 and Booster~T1 across locomotion, whole-body tracking, and loco-manipulation settings.
The main hardware benchmark contains five tasks.
\begin{description}[leftmargin=*, itemsep=2pt]
    \item[\textbf{G1 Slope Traversal.}]
    Unitree~G1 traverses two $15^\circ$ slopes on a narrow ramp of width approximately $0.8$\,m.
    The robot walks forward, turns $180^\circ$, and walks back over a total path length of approximately $20$\,m.
    A rollout is successful only if the robot completes the full out-and-back trajectory without stepping off the ramp.

    \item[\textbf{G1 Loco. + Payload.}]
    Unitree~G1 carries asymmetric grocery loads while following a sinusoidal path through three poles. The left hand carries a snack box, and the right hand carries a bag of oranges. The total load is approximately $3$\,kg, with the oranges being noticeably heavier and the snack box partially obstructing leg motion. We evaluate this task using normalized velocity-tracking error along the commanded path.

    \item[\textbf{G1 Kungfu + Soft Terrain.}]
    G1 tracks a Kungfu reference motion on deformable soft mats.
    Each hardware motion lasts approximately $15$\,s.
    We report normalized mean per-joint position error (MPJPE) and treat the rollout as failed if the robot falls before completing the motion.

    \item[\textbf{T1 Loco. + Payload.}]
    Booster~T1 tracks a circular path of radius $1$\,m while carrying an asymmetric $1$\,kg payload on the right arm.
    We evaluate this task using normalized velocity-tracking error around the commanded circle.

    \item[\textbf{T1 Basket Pulling.}]
    T1 pulls a plastic laundry basket loaded with laundry and weights, with a total mass of approximately $6$\,kg. The commanded motion is a backward straight-line track of approximately $6$ m. A rollout is considered successful if the robot pulls the basket across the finish line.

    \item[\textbf{G1 payload dancing.}]
    We include this as an additional qualitative hardware deployment rather than as part of the five-task quantitative benchmark.
    G1 tracks a dancing motion while carrying a front-mounted bag of approximately $3.2$\,kg for a $20$\,s motion.
    We use it to illustrate stability and motion completion after adaptation.
\end{description}

\paragraph{Command protocol and target data collection.}
All hardware evaluations use the same pre-scripted, open-loop command sequences for every method, without operator intervention.
Target rollouts for adaptation are collected with \idea-zs in the deployment condition; their commands need not match the evaluation commands.

\paragraph{Sim-to-sim tasks.}
For sim-to-sim evaluation, policies are trained in IsaacSim and evaluated under held-out MuJoCo target dynamics.
Commands are randomized for all sim-to-sim tasks.
\begin{description}[leftmargin=*, itemsep=2pt]
    \item[\textbf{G1 Slope Traversal.}]
    G1 follows randomized velocity commands on $20^\circ$ slopes.
    We evaluate this task using normalized velocity-tracking error.

    \item[\textbf{G1 Kungfu.}]
    G1 tracks the Kungfu reference motion under MuJoCo dynamics.
    The sequence is longer than the hardware deployment sequence, with a total duration of approximately $60$\,s.
    We evaluate this task using normalized mean per-joint position error (MPJPE).

    \item[\textbf{T1 Loco. + Payload.}]
    T1 follows randomized velocity commands with a $5$\,kg payload mounted on the torso.
    We evaluate this task using normalized velocity-tracking error.

    \item[\textbf{T1 Slope Traversal.}]
    T1 follows randomized velocity commands on $10^\circ$ slopes.
    We evaluate this task using normalized velocity-tracking error.

    \item[\textbf{T1 Falcon.}]
    T1 tracks commanded velocities while a constant $30$\,N downward force is applied to the right hand and no external force is applied to the left hand.
    We evaluate this force-adaptive locomotion task using normalized velocity-tracking error.
\end{description}
The payload sweep in \Cref{sec:experiments_operating_range} uses T1 Loco.\ + Payload with torso payloads of $0$, $5$, $10$, $15$, $20$, and $30$\,kg; survival rate is the fraction of commanded steps completed before failure.

\subsection{Observation Space}
\label[appendix]{app:observation_space}

\Cref{tab:obs_space} summarizes the observations available to the deployable student and to the privileged oracle.
The student observes only quantities measurable at deployment---proprioception, the task command, and a gait or motion reference---while the oracle additionally observes privileged simulator state such as base linear velocity, contact forces, terrain geometry, actuator parameters, and the sampled randomization parameters.
The two policy families (whole-body tracking and locomotion/loco-manipulation) share the same observation structure but differ in a few per-term scales, which we note in the table.
All policies use a history of $H{=}30$ stacked frames and a prediction horizon of $K{=}6$, as described in \Cref{sec:experiments}.

\input{tables/observation_space}

\subsection{Domain Randomization}
\label[appendix]{app:domain_randomization}

All source policies are trained in IsaacSim under domain randomization over dynamics, actuation, and perturbation parameters.
\Cref{tab:domain_randomization} lists the sampled ranges.
The whole-body tracking and locomotion families use the same randomization scheme with different magnitudes; whole-body motion tracking additionally randomizes the physical properties of the carried object.
These ranges define the source condition distribution $\srcdom$; the target conditions $\xi_{\mathrm{tgt}}$ evaluated in the main text are fixed deployment conditions that are not assumed to lie within these ranges.

\input{tables/domain_randomization_settings}

\subsection{Metrics and Normalization}
\label[appendix]{app:metrics}

We report task-appropriate metrics that separate binary completion from continuous tracking quality.
Completion tasks are reported as success rate.
Tracking tasks are reported as normalized errors, where lower is better and $1.0$ corresponds to the \tfdagger baseline error for the same task and evaluation condition.
For compact notation, we denote the corresponding \tfdagger baseline error by the superscript $\mathrm{TFD}$ in the formulas below.

\paragraph{Velocity tracking.}
For locomotion tasks, we compute linear and angular velocity RMSE over the evaluation window:
\begin{equation}
    E_{\mathrm{lin}}
    =
    \sqrt{\frac{1}{T}\sum_{t=1}^{T}
    \left\|
    v^{xy}_{t,\mathrm{cmd}} - v^{xy}_{t,\mathrm{base}}
    \right\|_2^2},
    \qquad
    E_{\mathrm{ang}}
    =
    \sqrt{\frac{1}{T}\sum_{t=1}^{T}
    \left(
    \omega^{z}_{t,\mathrm{cmd}}-\omega^{z}_{t,\mathrm{base}}
    \right)^2}.
\end{equation}
The reported normalized velocity error is
\begin{equation}
    \bar E_v =
    \frac{E_{\mathrm{lin}} + E_{\mathrm{ang}}}
    {E_{\mathrm{lin}}^{\mathrm{TFD}} + E_{\mathrm{ang}}^{\mathrm{TFD}}},
\end{equation}
computed per task under the same evaluation condition.

\paragraph{Whole-body tracking.}
For whole-body tracking, we report normalized mean per-joint position error,
\begin{equation}
    E_{\mathrm{mpjpe}}
    =
    \frac{1}{T |\mathcal{J}|}
    \sum_{t=1}^{T}\sum_{j \in \mathcal{J}}
    \left\|
    p_{t,j} - p^{\mathrm{ref}}_{t,j}
    \right\|_2,
    \qquad
    \bar E_{\mathrm{mpjpe}}
    =
    \frac{E_{\mathrm{mpjpe}}}{E_{\mathrm{mpjpe}}^{\mathrm{TFD}}},
\end{equation}
where $p_{t,j}$ and $p^{\mathrm{ref}}_{t,j}$ denote the deployed and reference joint or body-keypoint positions.
For hardware whole-body tracking, a trial is also treated as failed if the robot falls before completing the motion.

\paragraph{Success rate.}
Slope traversal succeeds if the robot completes the specified trajectory without falling or leaving the ramp.
Basket pulling succeeds if the basket crosses the finish line while the robot remains stable.
For whole-body tracking tasks, completion additionally requires the robot not to fall during the reference motion.

\section{\idea Implementation Details}
\label[appendix]{app:implementation}

This appendix records the implementation choices used for \idea in the main experiments.
We describe the deployable Planner--IDM architecture, source-domain pretraining, few-shot target adaptation, and the baselines used for comparison.
The end-to-end procedure is summarized in \Cref{alg:fada_procedure} of \Cref{sec:method}.

\subsection{Planner--IDM Architecture}
\label[appendix]{app:architecture}

The deployable policy is the factorized student $\pi^s=(P_\phi,I_\psi)$ from \Cref{sec:method}.
Both modules are transformer-based and operate on compact proprioceptive tokens rather than privileged simulator state.
The planner receives the recent proprioceptive observation history and command and predicts a future proprioceptive chunk $\hat Y_t^K$.
The IDM receives the same recent execution history together with a future chunk and predicts an action chunk $\hat U_t^K$.
In all main experiments, the history length is $H{=}30$ and the prediction horizon is $K{=}6$.
The planner and IDM encoder use 3-layer, 4-head transformers with hidden size 128, and the IDM decoder uses 2 transformer decoder layers with the same hidden size and number of heads.

The planner is implemented as a transformer encoder over observation-history and command-conditioned tokens.
In implementation, the planner head predicts a residual future chunk relative to the latest history observation.
We reconstruct the absolute future before passing it to the IDM,
\begin{equation}
    \hat Y_t^K = o_t + \Delta \hat Y_t^K,
\end{equation}
and use $\hat Y_t^K$ for the notation in the main text.

The IDM uses an encoder--decoder transformer interface.
Its history token at each timestep is formed by adding the observation embedding and the executed-action embedding, followed by positional encoding.
The planner similarly forms each history token by adding the observation embedding, a broadcast command embedding, and positional encoding.
Future-state tokens are formed by embedding each future observation in the $K$-step chunk and adding future positional encodings.
The IDM history encoder applies full self-attention over the history tokens.
The decoder takes all future-state tokens as target tokens; since no causal mask is used, future tokens have full self-attention over the entire predicted future chunk, and each future token cross-attends to all encoded history tokens.
The action head is applied to every decoded future token, producing the $K$-step action chunk in parallel.
The IDM output is always a $K$-step action chunk.
At deployment, the policy is executed in a receding-horizon manner: only the first action $\Pi_1(\hat U_t^K)$ is sent to the low-level controller, after which the history is updated and the planner and IDM are queried again.
This is why the main IDM objective supervises the first action of the chunk, while full-chunk action supervision is studied as an ablation.

\subsection{Source-Domain Training}
\label[appendix]{app:source_training}

As summarized in \Cref{sec:method_source_interface}, student-rollout states are relabeled by restoring simulator snapshots and rolling the final oracle forward for $K$ steps under the same command to obtain the oracle-shadow pair $(Y_{\mathrm{orac}}^K,U_{\mathrm{orac}}^K)$ and the relabeled action $a_t^\star$.
Here we record the additional implementation details.
Source-domain training first learns a privileged oracle policy $\pi^{\mathrm{o}}$ in IsaacSim with task rewards and domain randomization.
The oracle has access to privileged simulator information during training, but the deployable student does not.
The source student is trained with DAgger-style rollouts so that visited states can be paired with both rollout-policy trajectory data and privileged oracle labels.

Source buffers store both raw trajectory fields and oracle-shadow labels.
For rollouts generated by the final oracle, these two signals coincide.
For rollouts generated by the planner--IDM student policy, we additionally store the realized trajectory $(o_t,a_t,o_{t+1})$ alongside the oracle-shadow relabel described above.

Training routes these stored targets according to their role.
The planner updates use the oracle-shadow labels, so the predicted future is encouraged to be actionable by the final oracle through the stop-gradient IDM.
IDM updates use action-observation pairs that are causally matched.
Let $(Y_{\mathrm{traj}}^K,U_{\mathrm{traj}}^K)$ denote the realized trajectory pair and $(Y_{\mathrm{orac}}^K,U_{\mathrm{orac}}^K)$ denote the oracle-shadow pair.
The IDM training pair is selected as
\begin{equation}
    (Y_I^K,U_I^K)=
    \begin{cases}
    (Y_{\mathrm{traj}}^K,U_{\mathrm{traj}}^K), & \text{for trajectory-source batches},\\
    (Y_{\mathrm{orac}}^K,U_{\mathrm{orac}}^K), & \text{for oracle-source batches}.
    \end{cases}
\end{equation}
The latter is also valid inverse-dynamics supervision because the oracle-shadow chunk is generated by rolling the final oracle forward in the shadow simulator, so its actions and future observations form a physically realized causal pair under the oracle policy.
The IDM predicts an action chunk from $(\mathcal{O}_t^H,\mathcal{A}_t^H,Y_I^K)$, but the default loss supervises only $\Pi_1(U_I^K)$ to match receding-horizon deployment.
In the separate-pass training used in our main runs, source optimization alternates two passes within each training iteration: an IDM pass trained with full teacher forcing on these matched future-action pairs, and a planner pass that keeps IDM weights fixed while optimizing the planner through the IDM action loss.
Thus, oracle relabeling supplies intent labels, while trajectory fields preserve the realized dynamics needed by the IDM.

To broaden the IDM's coverage beyond near-optimal behavior, source data also includes suboptimal rollouts from intermediate oracle checkpoints.
The suboptimal data budget is set to twice the amount of optimal data, and $20$ intermediate oracle checkpoints are selected from the oracle training process.
This additional data is used only to improve the diversity of source-domain inverse-dynamics supervision; all source rollouts still carry oracle labels for planner supervision.

\subsection{Few-Shot Target Adaptation}
\label[appendix]{app:target_adaptation}

Target adaptation uses only ordinary target rollouts collected by executing the source-trained policy in the fixed deployment condition $\xi_{\mathrm{tgt}}$.
No target rewards, privileged target labels, simulator calibration, or source-domain replay are used during the target update.
Each target rollout is converted into the same window format as source IDM training:
$(\mathcal{O}_t^H,\mathcal{A}_t^H,Y_{t,\mathrm{exec}}^K,U_{t,\mathrm{exec}}^K)$.
The future observations and action chunks come from the same physical or simulated rollout window, so they encode how the target system actually responds to executed actions.

During adaptation, we freeze the Planner \(P_\phi\) and the pretrained IDM weights \(\psi\), and optimize only the LoRA parameters \(\Delta\psi\) inserted into the IDM.
We train these parameters with the action loss in \Cref{eq:adapt_loss}, using rank \(r=8\), scaling \(\alpha=16\), and dropout \(0.05\).
These values were selected from a small grid over \(r \in \{4,8,16\}\) and \(\alpha \in \{8,16,32\}\).
However, it should be noted that performance was not sensitive to these choices within the tested range.
Adaptation is performed once, offline, after the full target dataset is collected, for a fixed $800$ steps with learning rate $3{\times}10^{-4}$ and batch size $4096$; no online or iterative updates are used.
The resulting policy $(P_\phi,I_{\psi+\Delta\psi})$ keeps the same planner intent as the source policy but changes how that intent is translated into actions under $\xi_{\mathrm{tgt}}$.
This is the execution-only update tested throughout the main experiments.
Unlike source-domain pretraining, target adaptation has no oracle relabels.
Its supervision is entirely the paired target rollout data: the physically observed future $Y_{t,\mathrm{exec}}^K$ and the executed action chunk $U_{t,\mathrm{exec}}^K$.
The same first-action inverse-dynamics target $\Pi_1(U_{t,\mathrm{exec}}^K)$ is used, so the adapted IDM is aligned with the action actually executed under receding-horizon deployment.

\subsection{Baseline Implementations}
\label[appendix]{app:baselines}

We compare \idea against baselines that use the same source training pipeline, deployable observation space, and evaluation protocol whenever applicable.
The purpose of these baselines is to isolate whether few-shot target rollouts are most useful when they update the action-generation module rather than a monolithic student or a future-prediction objective.
Their interfaces are visualized in \Cref{fig:baselines} (main text).

\paragraph{\tfdagger.}
\tfdagger is a transformer teacher--student policy trained from the same source-domain DAgger data as \idea.
It maps recent observation-action history and the task command directly to the next action, without an explicit Planner--IDM factorization and without target-domain finetuning.

\paragraph{\copred-zs.}
\copred-zs is a zero-shot co-prediction baseline motivated by recent world-model and world-action-model approaches.
It uses a shared transformer backbone to jointly predict the next deployable observation and action from history and command inputs, and is evaluated directly in the target condition.

\paragraph{\copred-ft.}
\copred-ft uses the same target rollout budget as \idea but adapts the shared co-prediction backbone with a future-observation prediction objective on target windows.
This baseline tests whether target rollouts are sufficient when used to update a generic predictive representation, rather than the isolated IDM action-generation module.
\copred-ft uses the same LoRA configuration, target rollout budget, and number of adaptation steps as \idea.
Because \copred-ft degrades on all five sim-to-sim tasks (\Cref{tab:sim2sim_main}), we do not deploy it on hardware for safety reasons.

All target-adapted methods are compared using the same target rollout budget and evaluation commands.

\section{Additional Analysis and Ablations}
\label[appendix]{app:additional_analysis}

\subsection{Architecture Ablations}
\label[appendix]{app:design_ablations}

This ablation examines the role of transformer-based token processing in the Planner--IDM interface.
The experiment is run in the MuJoCo sim-to-sim setting on T1 Loco.\ + Payload, following the protocol of \Cref{app:tasks_conditions}, with $20$ randomized-command trials and results normalized column-wise by the first row.
LoRA vs.\ full IDM finetuning and the target rollout budget are studied in \Cref{sec:experiments_adaptation_design}.

\paragraph{Transformer vs. MLP modules.}
\Cref{tab:appendix_mlp_transformer} compares transformer modules against MLP replacements with the same input/output interface and a matched parameter budget on T1 Loco.\ + Payload.
The IsaacSim Eval column measures source-domain evaluation, while the FADA column reports post-adaptation MuJoCo performance.
The full transformer Planner--IDM gives the best source-domain evaluation and the best post-adaptation MuJoCo performance.
Replacing more of the architecture with MLP modules steadily weakens target adaptation, suggesting that attention over history and future tokens is important for learning a reusable dynamics-alignment interface.

\begin{table}[!htbp]
\centering
\caption{\textbf{Transformer vs. MLP Planner--IDM modules.} We report $\bar E_v\downarrow$ on T1 Loco. + Payload. IsaacSim Eval denotes source-domain evaluation, and FADA denotes post-adaptation MuJoCo evaluation. Codes denote planner / IDM encoder / IDM decoder, where T is transformer and M is MLP. MLP variants use the same inputs, outputs, and parameter budget as the transformer model.}
\label{tab:appendix_mlp_transformer}
\footnotesize
\setlength{\tabcolsep}{4.0pt}
\renewcommand{\arraystretch}{1.02}
\begin{tabular}{@{}lcc@{}}
\toprule
Code & IsaacSim Eval & FADA \\
\midrule
MMM & 1.0000$_{\pm0.0915}$ & 1.0000$_{\pm0.1235}$ \\
MTM & 0.9729$_{\pm0.0904}$ & 0.8345$_{\pm0.1129}$ \\
TTM & 0.9742$_{\pm0.0851}$ & 0.7664$_{\pm0.1210}$ \\
TTT (ours) & \goodnumber{0.9622$_{\pm0.0849}$} & \goodnumber{0.7056$_{\pm0.1150}$} \\
\bottomrule
\end{tabular}
\end{table}

\subsection{Same-Simulator Domain Shift Analysis}
\label[appendix]{app:same_sim_shift}

Same-simulator shifts isolate dynamics mismatch while avoiding additional confounders from simulator implementation or hardware deployment.
In this setting, we vary target payload or terrain conditions while keeping the simulator implementation, observation conventions, and reset procedures fixed.
\Cref{tab:appendix_same_sim} reports IsaacSim-to-IsaacSim transfer results under held-out target conditions.
Each policy is evaluated with $1024$ parallel IsaacSim environments, enabling a statistically more reliable comparison than smaller-scale hardware trials.
Since source and target share the same simulator implementation, the domain gap is smaller than in MuJoCo or real-world deployment.
Nevertheless, \idea consistently improves over zero-shot transfer across payload and terrain shifts.

\begin{table}[!htbp]
\centering
\caption{\textbf{IsaacSim-to-IsaacSim held-out domain shifts.} We report $\bar E_v\downarrow$ using $1024$ parallel IsaacSim evaluation environments. Even when source and target share the same simulator implementation, \idea improves over zero-shot transfer on held-out dynamics.}
\label{tab:appendix_same_sim}
\footnotesize
\setlength{\tabcolsep}{3.0pt}
\renewcommand{\arraystretch}{1.02}
\begin{tabular}{@{}lccc@{}}
\toprule
Method & \makecell{G1 Loco.\\+ Payload} & \makecell{T1 Slope\\Traversal} & \makecell{T1 Loco.\\+ Payload} \\
\midrule
\tfdagger & 1.0000$_{\pm0.0444}$ & 1.0000$_{\pm0.0418}$ & 1.0000$_{\pm0.0278}$ \\
\idea-zs & 1.0501$_{\pm0.0234}$ & 0.9906$_{\pm0.0351}$ & 1.0177$_{\pm0.0189}$ \\
\idea (ours) & \goodnumber{0.8514$_{\pm0.0282}$} & \goodnumber{0.9461$_{\pm0.0243}$} & \goodnumber{0.9380$_{\pm0.0215}$} \\
\bottomrule
\end{tabular}
\end{table}

\section{Additional Sim-to-Real Results}
\label[appendix]{app:additional_results}
\label[appendix]{app:sim2real_extra}

\begin{figure*}[!t]
    \centering
    \includegraphics[width=\linewidth]{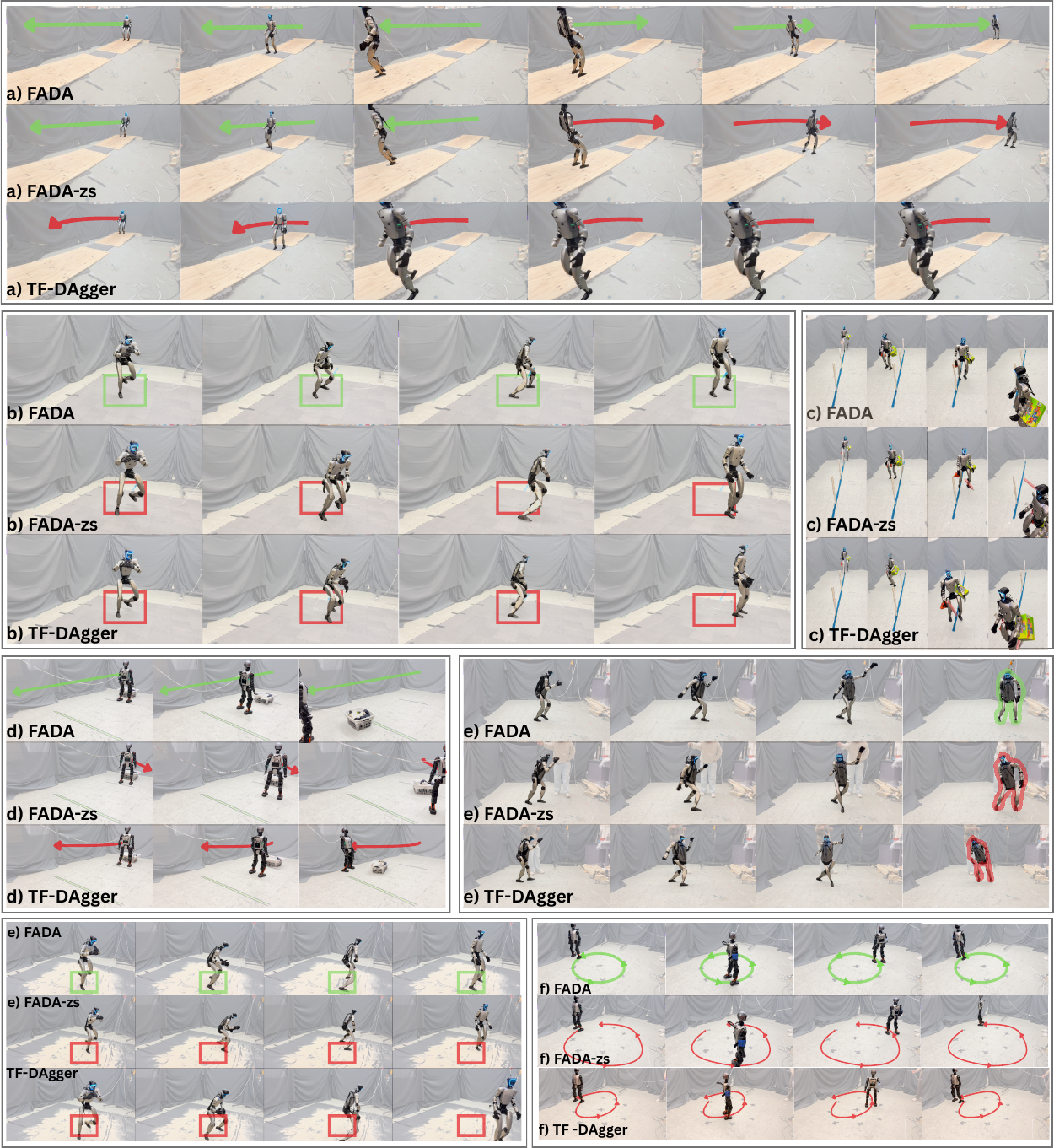}
    \caption{
    \textbf{Additional qualitative sim-to-real deployments.}
    We compare \idea, \idea-zs, and \tfdagger across seven hardware settings:
    \textbf{(a)} G1 slope walking,
    \textbf{(b)} G1 Kungfu tracking on soft mats,
    \textbf{(c)} G1 grocery carrying through poles,
    \textbf{(d)} T1 laundry-basket dragging,
    \textbf{(e)} G1 dance with payload,
    \textbf{(f)} G1 Kungfu tracking on sand, and
    \textbf{(g)} T1 circle walking with payload.
    Green annotations indicate successful tracking, stable regions, or completed paths; red annotations indicate drift, instability, or task failure. Across tasks, IDM adaptation improves execution under terrain, payload, and contact mismatch.
    }
    \label{fig:sim2real}
\end{figure*}

In addition to the hardware results reported in the main text, we provide qualitative rollouts across a broader set of deployment conditions in \Cref{fig:sim2real}.
These trials compare \idea against the zero-shot variant \idea-zs and the \tfdagger baseline on contact-rich terrain, payload transport, and whole-body tracking tasks.
Across settings, the same trend appears: zero-shot policies often produce partially correct motion but lose tracking under the target dynamics, while IDM adaptation improves the execution needed to realize the commanded behavior.

\Cref{fig:sim2real} shows seven representative examples.
On G1 slope walking, \idea maintains progress along the plank, whereas the baselines drift or step off the traversable region.
On soft-mat and sand-terrain Kungfu tracking, adaptation improves posture stability and keeps the robot closer to the commanded tracking region despite deformable contact.
For payload tasks, including grocery carrying, dance with payload, and T1 circle walking with payload, \idea better compensates for load-induced changes in balance and body velocity, reducing drift and improving path following.
In the T1 laundry-basket dragging task, the adapted policy maintains a straighter pulling direction and reaches the finish region, while the baselines fail to sustain the required contact-rich motion.

These examples highlight two recurring failure modes of zero-shot transfer.
First, under terrain or contact mismatch, the policy may still initiate the correct skill but accumulate foot-placement and posture errors over time.
Second, under payload or dragging forces, the planned motion remains reasonable, but the action mapping is no longer sufficient to realize the intended body motion.
IDM finetuning addresses these failures by recalibrating the plan-to-action map from target rollouts, without changing the planner or requiring task-specific rewards.

%% file: tables/observation_space.tex
\begin{table}[!htbp]
\centering
\caption{\textbf{Observation space.} The deployable student observes proprioception, the task command, and a gait/motion reference; the oracle additionally observes the privileged simulator state listed in the bottom block. The listed scale is the multiplicative factor applied to each term before concatenation; the privileged terms are compact state \emph{bundles} that group several quantities (e.g.\ contact forces, terrain heights, actuator parameters) and are internally normalized to comparable magnitudes within the observation function, so an outer scale of $1.0$ passes them through unchanged. Scales are the locomotion-family values; whole-body-tracking (WBT) values are noted in parentheses where they differ.}
\label{tab:obs_space}
\small
\renewcommand{\arraystretch}{1.15}
\begin{tabular}{llc}
\toprule
\textbf{Symbol} & \textbf{Description} & \textbf{Scale} \\
\midrule
\multicolumn{3}{c}{\textit{Common student observations (all tasks)}} \\
\midrule
$\boldsymbol{\omega}^{\mathrm{base}}_t$ & Base angular velocity & $0.25$~{\tiny(WBT $1.0$)} \\
$\mathbf{g}_t$ & Projected gravity vector & $1.0$ \\
$\mathbf{q}_t$ & Joint positions & $1.0$ \\
$\dot{\mathbf{q}}_t$ & Joint velocities & $0.05$~{\tiny(WBT $1.0$)} \\
$\mathbf{a}_{t-1}$ & Previous action & $1.0$ \\
\midrule
\multicolumn{3}{c}{\textit{Task-specific student observations}} \\
\midrule
$v^{x}_{\mathrm{cmd}},\,v^{y}_{\mathrm{cmd}},\,\omega^{z}_{\mathrm{cmd}}$ & Velocity command (locomotion) & $1.0$ \\
$\sin\Phi,\,\cos\Phi$ & Gait phase (locomotion) & $1.0$ \\
$q^{\mathrm{ref}}_{\mathrm{upper}}$ & Upper-body reference pose (loco-manip.) & $1.0$ \\
$h_{\mathrm{cmd}}$ & Base-height command (loco-manip.) & $2.0$ \\
$\mathbf{m}_t$ & Motion command (WBT) & $1.0$ \\
$R^{\mathrm{ref}}_{b}$ & Reference orientation in base frame (WBT) & $1.0$ \\
\midrule
\multicolumn{3}{c}{\textit{Privileged observations (oracle only)}} \\
\midrule
$\mathbf{v}^{\mathrm{base}}_t$ & Base linear velocity & $2.0$~{\tiny(WBT $1.0$)} \\
$\mathbf{c}_t$ & Per-body contact forces and binary contacts & $1.0$ \\
$\mathbf{h}_t$ & Local terrain height profile and root clearance & $1.0$ \\
$\boldsymbol{\theta}^{\mathrm{act}}_t$ & Actuator state (PD-gain scales, normalized torques) & $1.0$ \\
$\boldsymbol{\xi}_t$ & Sampled domain-randomization parameters & $1.0$ \\
$\mathbf{p}^{\mathrm{ref}}_{b},R^{\mathrm{ref}}_{b}$ & Reference body poses in base frame (WBT) & $1.0$ \\
$\mathbf{f}^{\mathrm{ee}}_t$ & End-effector applied force (loco-manip.) & $0.1$ \\
\bottomrule
\end{tabular}
\end{table}

%% file: tables/domain_randomization_settings.tex
\begin{table}[t]
\centering
\caption{\textbf{Domain randomization ranges} applied during source-domain training in IsaacSim. $\mathcal{U}(a,b)$ denotes a uniform range; $\times$ denotes a multiplicative scale on the nominal value. Whole-body tracking and the locomotion family use the same randomization scheme with different magnitudes; the carried-object terms apply only to whole-body motion tracking.}
\label{tab:domain_randomization}
\small
\setlength{\tabcolsep}{6pt}
\renewcommand{\arraystretch}{1.15}
\begin{tabular}{lcc}
\toprule
\textbf{Parameter} & \textbf{Whole-Body Tracking} & \textbf{Locomotion \& Loco-Manip.} \\
\midrule
\multicolumn{3}{c}{\textit{Dynamics randomization}} \\
\midrule
Ground friction              & $\mathcal{U}(0.1,2.0)$              & $\mathcal{U}(0.1,2.0)$ \\
Restitution                  & $\mathcal{U}(0.0,0.75)$            & -- \\
Base CoM shift ($x$)         & $\mathcal{U}(-0.025,0.025)$\,m     & $\mathcal{U}(-0.15,0.15)$\,m \\
Base CoM shift ($y,z$)       & $\mathcal{U}(-0.05,0.05)$\,m       & $\mathcal{U}(-0.15,0.15)$\,m \\
Added base mass              & $\mathcal{U}(-2.0,4.0)$\,kg        & $\mathcal{U}(-3.0,6.0)$\,kg \\
Link mass scale              & $\mathcal{U}(0.9,1.1)\times$       & $\mathcal{U}(0.8,1.3)\times$ \\
DoF position bias            & $\mathcal{U}(-0.01,0.01)$\,rad     & $\mathcal{U}(-0.05,0.05)$\,rad \\
\midrule
\multicolumn{3}{c}{\textit{Actuation and control}} \\
\midrule
$K_p$ scale                  & $\mathcal{U}(0.9,1.1)\times$       & $\mathcal{U}(0.8,1.2)\times$ \\
$K_d$ scale                  & $\mathcal{U}(0.9,1.1)\times$       & $\mathcal{U}(0.8,1.2)\times$ \\
Torque RFI                   & $0.1\,\tau_{\max}$                 & $0.1\,\tau_{\max}$ \\
Control delay                & $\{0,1\}$ steps                    & $\{0,1\}$ steps \\
\midrule
\multicolumn{3}{c}{\textit{Perturbations}} \\
\midrule
Push interval                & $\mathcal{U}(1,3)$\,s              & $\mathcal{U}(5,10)$\,s \\
Max push velocity            & $0.5$\,m/s, $0.52$\,rad/s          & $\mathcal{U}(0.1,1.5)$ \\
\midrule
\multicolumn{3}{c}{\textit{Carried-object randomization (whole-body tracking)}} \\
\midrule
Object friction              & \multicolumn{2}{c}{$\mathcal{U}(0.1,0.6)$} \\
Object restitution           & \multicolumn{2}{c}{$\mathcal{U}(0.0,1.0)$} \\
Object mass scale            & \multicolumn{2}{c}{$\mathcal{U}(1.0,4.0)\times$} \\
Object inertia ($I_{xx}$) scale & \multicolumn{2}{c}{$\mathcal{U}(0.5,1.5)\times$} \\
\bottomrule
\end{tabular}
\end{table}